\documentclass[times, review, 10pt]{elsarticle}

\usepackage{amssymb}
\usepackage{amsmath}
\usepackage{amsthm}

\usepackage{lineno}
\usepackage{multirow}
\usepackage{multicol}
\usepackage{algorithm}
\usepackage{algorithmic} 
\usepackage{xspace}
\usepackage{subcaption}
\usepackage{booktabs}
\usepackage{colortbl}
\usepackage{xcolor}
\usepackage{pifont}
\usepackage{enumitem}
\usepackage{url}
\usepackage[most]{tcolorbox}
\usepackage{makecell}
\usepackage{placeins}
\let\oldsubsection\subsection
\renewcommand{\subsection}{\FloatBarrier\oldsubsection}

\newcommand{\rev}[1]{{\color{black}#1}}

\newtcolorbox{promptbox}[1]{
    enhanced,
    colback=gray!5,
    colframe=black,
    coltitle=white,
    colbacktitle=black,
    fonttitle=\bfseries,
    title=#1,
    boxrule=1.5pt,
    arc=3pt,
    left=10pt, right=10pt, top=8pt, bottom=8pt,
    fontupper=\rmfamily,
}

\newtheorem{definition}{Definition}
\newtheorem{proposition}{Proposition}

\newcommand{\ours}{LaPA\textsuperscript{2}\xspace}

\journal{Pattern Recognition}

\begin{document}

\begin{frontmatter}

\title{LaPA\textsuperscript{2}: Length-Aware Prefix and Prompt Attention Augmentation for Long-Form Controllable Text Generation}

\author[aff1,aff2]{Jiabing Yang}
\ead{yangjiabing2025@ia.ac.cn}
\author[aff1,aff2]{Yixiang Chen}
\ead{yixiang.chen@cripac.ia.ac.cn}
\author[aff3]{Zichen Wen}
\ead{zichen.wen@outlook.com}
\author[aff4]{Chenhang Cui}
\ead{chenhangcui@gmail.com}
\author[aff1,aff2]{Peiyan Li}
\ead{peiyan.li@cripac.ia.ac.cn}
\author[aff1,aff2]{Yuan Xu}
\ead{yuan.xu@nlpr.ia.ac.cn}
\author[aff1,aff2]{Bowen Fang}
\ead{bwn.fang@gmail.com}
\author[aff1,aff2]{Tao Yu}
\ead{yutao2025@ia.ac.cn}
\author[aff5]{Ruikang Lin}
\ead{linrk.proton@gmail.com}
\author[aff1,aff2]{Yan Huang\corref{cor1}}
\ead{yhuang@nlpr.ia.ac.cn}
\author[aff1,aff2]{Liang Wang}
\ead{wangliang@nlpr.ia.ac.cn}
\cortext[cor1]{Corresponding author.}

\affiliation[aff1]{organization={School of Artificial Intelligence},
            addressline={University of Chinese Academy of Sciences},
            city={Beijing},
            country={China}}
\affiliation[aff2]{organization={New Laboratory of Pattern Recognition (NLPR)},
            addressline={Institute of Automation, Chinese Academy of Sciences (CASIA)},
            city={Beijing},
            country={China}}
\affiliation[aff3]{organization={School of Artificial Intelligence},
            addressline={Shanghai Jiao Tong University},
            city={Shanghai},
            country={China}}
\affiliation[aff4]{organization={School of Computing},
            addressline={National University of Singapore},
            country={Singapore}}
\affiliation[aff5]{organization={School of Computer Science and Engineering},
            addressline={University of Electronic Science and Technology of China},
            city={Chengdu},
            country={China}}

\begin{abstract}
Prefix-based methods have emerged as a promising paradigm for Controllable Text Generation (CTG) due to their parameter efficiency. However, while effective in short sequences, their controllability tends to diminish as the generated sequence grows. In this paper, we identify \textbf{Attention Dilution} as a key factor behind this phenomenon: as the sequence length increases, the attention allocated to the control signal naturally decays due to the softmax mechanism, leading to a ``fading'' control effect. To address this, we propose \textbf{\ours} (\textbf{L}ength-\textbf{a}ware \textbf{P}refix and \textbf{P}rompt \textbf{A}ttention \textbf{A}ugmentation), a training-free and model-agnostic framework designed to sustain robust control in long contexts. Specifically, \ours employs \textbf{Length-Aware Logarithmic Scaling} to dynamically amplify prefix attention weights, mathematically counteracting the dilution effect, while an optional \textbf{Contextual Anchor Reinforcement} applies synchronized augmentation to prompt tokens, preserving semantic coherence when strong attribute control risks overshadowing the original prompt. \ours is versatile, supporting both soft prefixes (continuous embeddings) and hard prefixes (discrete instructions). Experiments on multiple CTG tasks demonstrate that \ours consistently improves the performance of various prefix-based methods in long-form settings, leading to superior attribute controllability while preserving content relevance and fluency. Our code and data are publicly available at \url{https://github.com/jiabingyang01/LaPA2}.
\end{abstract}

\begin{graphicalabstract}
\includegraphics[width=\textwidth]{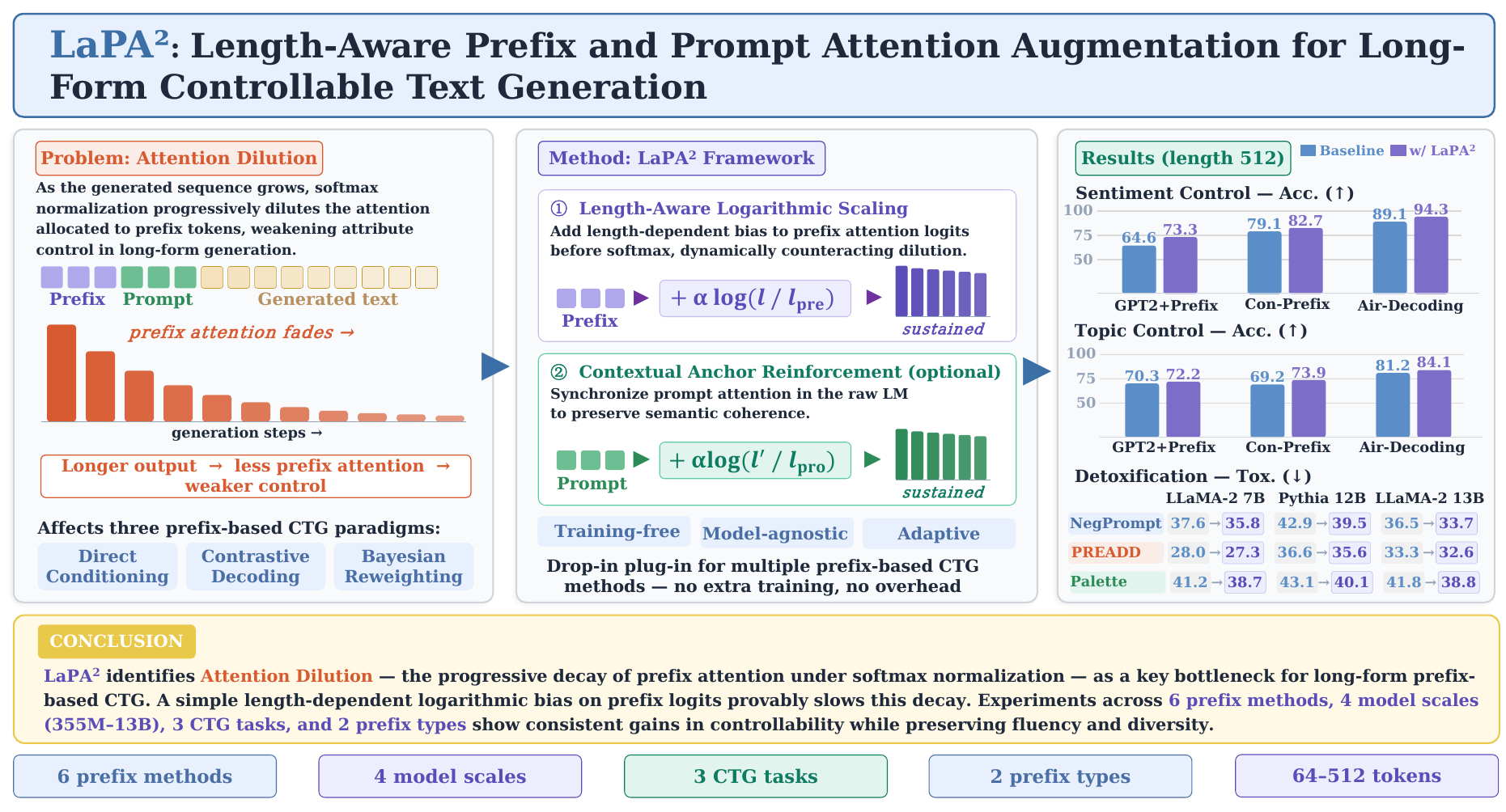}
\end{graphicalabstract}

\begin{highlights}
\item Prefix attention dilution identified as a key factor in long-form control loss.
\item Formal proof that prefix attention decays inversely with sequence length.
\item Length-Aware Scaling provably slows the prefix attention decay rate.
\item Optional Anchor Reinforcement preserves prompt coherence.
\item Consistent gains across \rev{six} prefix methods, four model scales, three tasks.
\end{highlights}

\begin{keyword}
Controllable Text Generation \sep Attention Dilution \sep Prefix-Based Methods \sep Long-Form Generation
\end{keyword}

\end{frontmatter}


\section{Introduction}

Controllable Text Generation (CTG) aims to steer language models toward desired attributes such as sentiment, topic, or safety \citep{Survey, zhang2023survey}. Various approaches have been proposed, ranging from full model retraining \citep{Ctrl, CoCon} and reinforcement learning \citep{li2024reinforcement, zeng2024token} to parameter-efficient fine-tuning \citep{Discup, Con-Prefix} and decoding-time intervention \citep{DExperts, FUDGE, GeDi, Air-Decoding}. Among these, prefix-based methods are especially appealing: by prepending attribute-specific tokens, either learned embeddings \citep{Prefix-Tuning, Con-Prefix, Air-Decoding} or natural language instructions \citep{PREADD, dekoninckcontrolled}, to the model input, they achieve effective control without modifying model weights.

\begin{figure}[!t]
    \centering
    \includegraphics[width=0.7\linewidth]{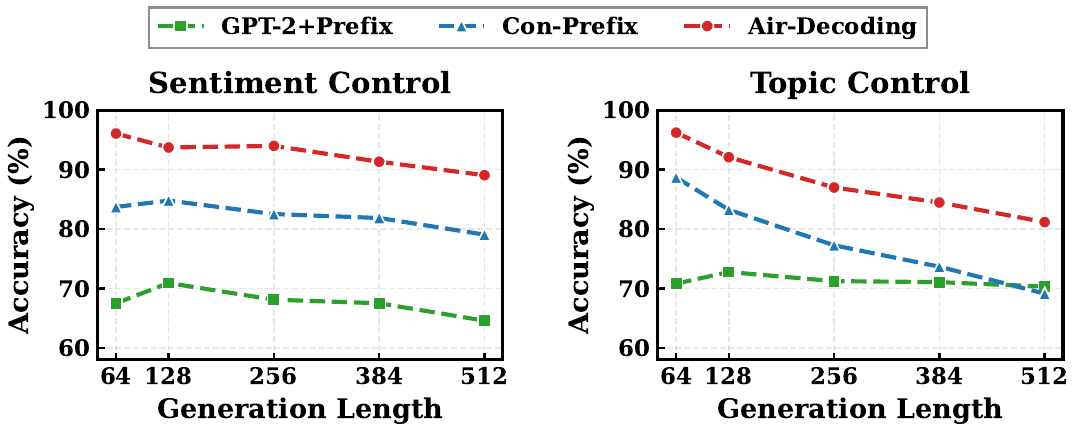}
    \caption{Attribute accuracy of three prefix-based methods across generation lengths on sentiment and topic control (GPT-2 Medium). Controllability declines as sequences grow longer.}
    \label{fig:decay}
\end{figure}

However, existing evaluations of prefix-based methods have been largely confined to short-form generation, typically 20--50 tokens \citep{PPLM, DExperts, Air-Decoding, FreeCtrl}. This narrow scope masks a practical weakness: as shown in Figure~\ref{fig:decay}, the controllability of prefix-based methods \emph{tends to degrade} as the generated sequence grows. \rev{While recent architectural solutions such as Non-Residual Prompting \citep{NRP} and the Residual Memory Transformer \citep{RMT} mitigate this decay, they require additional trainable modules and dedicated pre-training.} This limits their applicability to longer-output scenarios such as story or document-level generation.

Inspired by findings that visual attention decay leads to degraded generation quality in multimodal domains \citep{tang2025seeing, IKOD, yang2026uaor}, we hypothesize that an analogous \textbf{Attention Dilution} underlies the degradation in prefix-based CTG: under softmax normalization, the expected attention weight allocated to prefix tokens monotonically decreases as the total sequence length increases. We formalize this phenomenon and show that, under a uniform logit assumption, expected prefix attention decays at a rate of $O(l^{-1})$, where $l$ is the total sequence length. Empirical evidence strongly supports this hypothesis, providing a principled explanation for the observed control loss in prefix-based CTG methods.

Based on this analysis, we propose \textbf{L}ength-\textbf{a}ware \textbf{P}refix and \textbf{P}rompt \textbf{A}ttention \textbf{A}ugmentation (\textbf{\ours}), a training-free framework that counteracts attention dilution with two mechanisms. \textbf{Length-Aware Logarithmic Scaling} adds a length-dependent bias to prefix attention logits, reducing the decay rate from $O(l^{-1})$ to $O(l^{\alpha-1})$ (Proposition~\ref{prop:correction}). An optional \textbf{Contextual Anchor Reinforcement} applies analogous augmentation to prompt tokens, preserving semantic coherence for methods where strong attribute control may overshadow the original prompt. Since the augmentation operates on attention logits, \ours is agnostic to the prefix type and can serve as a drop-in module for methods using either soft or hard prefixes. \rev{Unlike retraining-based \citep{Ctrl, CoCon} or auxiliary-module approaches \citep{NRP, RMT}, \ours introduces no learnable parameters and no additional forward passes.} We evaluate its effectiveness across \rev{six} prefix-based methods, four models (355M--13B), three CTG tasks, and generation lengths from 64 to 512 tokens.
Our main contributions are as follows:
\begin{enumerate}[leftmargin=*]
    \item We identify \textbf{Attention Dilution} as a key factor in the long-form degradation of prefix-based CTG methods, and formally show that, under a uniform logit assumption, prefix attention decays as $O(l^{-1})$.
    \item We propose \ours, a \rev{\textbf{training-free, zero-parameter, and model-agnostic}} framework that employs \textbf{Length-Aware Logarithmic Scaling} to reduce the prefix attention decay rate, with an optional \textbf{Contextual Anchor Reinforcement} to preserve semantic coherence.
    \item Extensive experiments across \rev{six} prefix-based methods, four model scales, and three CTG tasks demonstrate that \ours consistently improves controllability in long-form settings while preserving fluency and content relevance.
\end{enumerate}

\section{Related Work}

\subsection{Retraining-Based CTG}
Early CTG methods modify model architectures or parameters using attribute-specific data. Keskar et al.\ \citep{Ctrl} train a 1.63B-parameter conditional Transformer with 55 control codes. CoCon \citep{CoCon} injects control condition embeddings into hidden states via self-supervised blocks. POINTER \citep{POINTER} is pre-trained with a progressive insertion-based objective on 12GB of Wikipedia and fine-tuned for hard-constrained generation. Director \citep{Director} introduces a generator-classifier architecture that refines each token's output by combining language model and classifier heads. While effective, these methods incur substantial computational costs, motivating lightweight alternatives.

\subsection{Prefix/Prompt-Based CTG}
As language models scale, parameter-efficient methods have become widely adopted. Prefix-Tuning \citep{Prefix-Tuning} prepends trainable continuous embeddings to each Transformer layer, and Prompt-Tuning \citep{Prompt-Tuning} inserts them only at the input layer. Qian et al.\ \citep{Con-Prefix} train contrastive prefixes jointly for opposing attributes. 
PREADD \citep{PREADD} and Model Arithmetic \citep{dekoninckcontrolled} instead use natural language instructions as hard prefixes, avoiding the need for any parameter updates. \rev{Palette \citep{Palette} further improves multi-attribute combination by modeling attribute overlaps via conditional mutual information minimization over prompt-conditioned distributions.} \rev{Prompt-based control has also been explored in multimodal generation: QPDC \citep{QPDC} uses question-driven prompts with uncertainty-aware fusion for controllable video captioning, and RSFD \citep{RSFD} enhances low-frequency token semantics to improve long-sequence consistency. To address control signal decay over long sequences, Non-Residual Prompting \citep{NRP} introduces position-invariant key-values via a non-residual attention stream, but requires cloning the base CLM and multi-phase pre-training. The Residual Memory Transformer \citep{RMT} uses a cross-attention plugin to apply control conditions uniformly at every step, but still requires pre-training and fine-tuning. In contrast, \ours counteracts the decay with a training-free logit bias applicable to various prefix-based methods.}

\subsection{Decoding-Time Intervention}
These methods modify the output distribution during inference. PPLM \citep{PPLM} uses gradients from an attribute classifier to steer hidden states, while FUDGE \citep{FUDGE} applies Bayesian factorization to adjust token probabilities. DExperts \citep{DExperts} combines expert and anti-expert models via contrastive decoding, and ROSE \citep{zhong2024rose} boosts safety through reverse prompt contrastive decoding. GeDi \citep{GeDi} and Air-Decoding \citep{Air-Decoding} use class-conditional language models with control code prefixes to guide generation via Bayes' rule. Other directions include distributional constraints \citep{khalifa2021distributional} and pragmatic reasoning \citep{RSA-Control}. Many of these methods internally rely on prefix-conditional language models, making them susceptible to the attention dilution problem we identify.

\section{Understanding Attention Dilution}
\label{sec:dilution}

\subsection{Preliminaries: Prefix-Based CTG}
\label{sec:preliminary}

Given a prompt $x_{1:T-1}$ and a target attribute $a$ (e.g., positive sentiment), CTG aims to generate a continuation $x_{T:N}$ aligned with $a$:
\begin{equation}
P(x_{T:N}|x_{1:T-1},a) = \prod_{t=T}^N P(x_t|x_{<t},a)
\label{eq:ctg}
\end{equation}
Prefix-based methods achieve this by prepending attribute-specific tokens to the model input. We distinguish two types: \textbf{soft prefixes}, which are continuous embeddings learned from attribute-specific data \citep{Prefix-Tuning, Con-Prefix}, and \textbf{hard prefixes}, which are discrete natural language instructions (e.g., ``\textit{A positive text:}'') \citep{PREADD, dekoninckcontrolled}. Both types influence generation through the same attention mechanism. Let $P_{\theta_a}(x_t|x_{<t})$ denote the output distribution when prefix $\theta_a$ (encoding attribute $a$) is prepended. Existing methods differ in how $P_{\theta_a}$ is used to derive the controlled distribution $P(x_t|x_{<t},a)$. We identify three paradigms:

\noindent \textbf{(I) Direct Prefix Conditioning.} The simplest approach uses the prefix-conditioned output directly: $P(x_t|x_{<t},a) = P_{\theta_a}(x_t|x_{<t})$. This includes Prefix-Tuning \citep{Prefix-Tuning}, Contrastive Prefixes \citep{Con-Prefix}, and NegPrompt \citep{PREADD}, which prepends a target attribute prefix (e.g., ``\textit{Very positive:}'').

\noindent \textbf{(II) Contrastive Decoding.} Rather than using the prefix-conditioned output directly, these methods modulate generation by contrasting output distributions obtained with and without the attribute prefix. The log-probability difference induced by the prefix
 is:
\begin{equation}
d_t = \log P_{\theta_a}(x_t|x_{<t}) - \log P(x_t|x_{<t})
\label{eq:logdiff}
\end{equation}
The controlled logit is then modeled as $\log P(x_t|x_{<t}) + \lambda \cdot d_t$, where $\lambda$ modulates the control strength. Converting back to probability space yields:
\begin{equation}
P(x_t|x_{<t},a) \propto P_{\theta_a}(x_t|x_{<t})^{\lambda} \cdot P(x_t|x_{<t})^{1-\lambda}
\label{eq:contrastive}
\end{equation}
Setting $\lambda > 1$ amplifies the prefix's effect beyond direct conditioning; setting $\lambda < 0$ enables \emph{negative control} against the prefix attribute (e.g., using a toxic prefix with $\lambda < 0$ for detoxification). PREADD \citep{PREADD} adopts this formulation with hard prefixes, and ROSE \citep{zhong2024rose} applies reverse prompt contrastive decoding for safety enhancement.

\noindent \textbf{(III) Bayesian Reweighting.} These methods factor the controlled distribution through Bayes' rule, using prefix-conditional models to estimate an attribute classifier. Following Yang and Klein \citep{FUDGE}, the controlled distribution can be written as:
\begin{equation}
P(x_t|x_{<t},a) \propto P(a|x_{1:t})^{\omega} \cdot P(x_t|x_{<t}), \quad t \geq T
\label{eq:bayes}
\end{equation}
where $P(a|x_{1:t})$ is an attribute classifier estimated from prefix-conditional models via Bayes' rule:
\begin{equation}
P(a|x_{1:t})=\frac{\prod_{j=T}^t P_{\theta_a}(x_j|x_{<j})}{\sum_{a^{'}\in\{a,\bar{a}\}}\prod_{j=T}^t P_{\theta_{a^{'}}}(x_j|x_{<j})}
\label{eq:classifier}
\end{equation}
assuming equal class priors, i.e., $P(a)=P(\bar{a})=1/2$. Following Air-Decoding \citep{Air-Decoding}, we detail the derivation of the above two equations below. For Eq.~\ref{eq:bayes}, starting from the definition of conditional probability and applying Bayes' rule:
\begin{align}
    P(x_t|x_{1:t-1},a)
    &= \frac{P(x_{1:t},a)}{P(x_{1:t-1},a)} \nonumber\\
    &= \frac{P(a|x_{1:t})\,P(x_{1:t})}{P(a|x_{1:t-1})\,P(x_{1:t-1})} \nonumber\\
    &= \frac{P(a|x_{1:t})}{P(a|x_{1:t-1})} \cdot P(x_t|x_{1:t-1}) \nonumber\\
    &\propto P(a|x_{1:t}) \cdot P(x_t|x_{1:t-1})
    \label{eq:bayes_deriv}
\end{align}
where the last step treats $P(a|x_{1:t-1})$ as a constant with respect to $x_t$. For Eq.~\ref{eq:classifier}, the classifier $P(a|x_{1:t})$ is obtained by applying Bayes' rule and the autoregressive factorization of the prefix-conditional model:
\begin{align}
    P(a|x_{1:t})
    &= \frac{P(a)\,P_{\theta_a}(x_{1:t})}{\sum_{a^{'}\in\{a,\bar{a}\}} P(a^{'})\,P_{\theta_{a^{'}}}(x_{1:t})} \nonumber\\
    &= \frac{P(a)\,\prod_{j=T}^{t} P_{\theta_a}(x_j|x_{<j})}{\sum_{a^{'}\in\{a,\bar{a}\}} P(a^{'})\,\prod_{j=T}^{t} P_{\theta_{a^{'}}}(x_j|x_{<j})}
    \label{eq:classifier_deriv}
\end{align}

GeDi \citep{GeDi} trains class-conditional language models with control code prefixes to compute $P(a|x_{1:t})$ and guide generation. Air-Decoding \citep{Air-Decoding} further reconstructs attribute distributions from the class-conditional outputs. RSA-Control \citep{RSA-Control} extends this paradigm by incorporating pragmatic reasoning into the reweighting process. \rev{Palette \citep{Palette} generalizes it via the Law of Total Probability to model attribute overlaps.}

Despite their architectural differences, all three paradigms share a common bottleneck: the prefix $\theta_a$ influences generation exclusively through the \emph{attention mechanism} in the Transformer. As we show next, this shared dependency makes all prefix-based methods susceptible to the same limitation.

\subsection{Empirical Evidence of Attention Dilution}
\label{sec:empirical}

\begin{figure}[!t]
    \centering
    \includegraphics[width=0.7\linewidth]{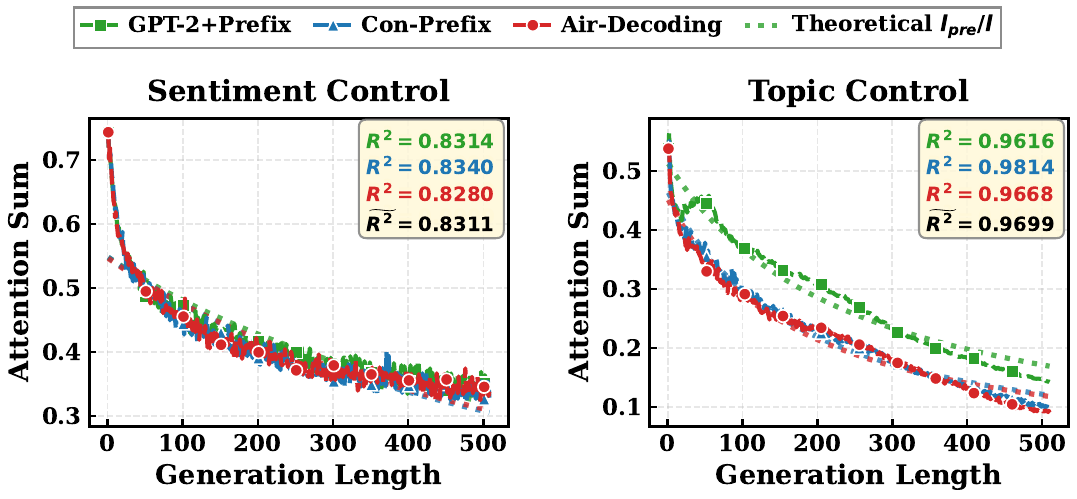}
    \caption{Prefix attention sum across generation steps. Dotted lines show fitted inverse proportional curves $C/(t+D)$, with high $R^2$ suggesting the theoretical $O(l^{-1})$ decay rate (Proposition~\ref{prop:decay}).}
    \label{fig:dilution}
\end{figure}

As shown in Figure~\ref{fig:decay}, attribute accuracy declines with generation length for all three prefix-based methods (GPT-2+Prefix, Con-Prefix, Air-Decoding) on both sentiment and topic control, where sentiment experiments use hard prefixes and topic experiments use soft prefixes, showing that the degradation occurs regardless of prefix type. Figure~\ref{fig:dilution} reveals a strong correlate: prefix attention sum decays monotonically as sequences grow. Moreover, we fit an inverse proportional curve $C/(t+D)$ to each method's attention trajectory, where $t$ denotes the generation step and the offset $D$ absorbs the fixed prefix and prompt lengths; the high average $R^2$ confirms that the empirical decay closely follows the $O(l^{-1})$ trend predicted by our analysis (\S\ref{sec:formal}), indicating that attention dilution affects prefix-based methods broadly.

\begin{table}[t]
\caption{Attribute relevance (\%) of first-half vs.\ second-half output segments across generation lengths under Air-Decoding. The second half consistently shows lower relevance, confirming progressive control loss.}
\label{tab:half}
\small
\centering
\resizebox{0.7\linewidth}{!}{
\begin{tabular}{ccccccc}
\toprule
& \multicolumn{3}{c}{\textbf{Sentiment Control}} & \multicolumn{3}{c}{\textbf{Topic Control}} \\
\cmidrule(lr){2-4} \cmidrule(lr){5-7}
\textbf{Length} & \textbf{First} & \textbf{Second} & \textbf{Full} & \textbf{First} & \textbf{Second} & \textbf{Full} \\
\midrule
256 & \textbf{94.33} & 91.83 & 94.00 & \textbf{93.45} & 86.05 & 87.00 \\
512 & \textbf{92.47} & 74.93 & 89.07 & \textbf{88.23} & 74.90 & 81.18 \\
\bottomrule
\end{tabular}}
\end{table}

To further localize the degradation, we split Air-Decoding's outputs into two halves and assess attribute relevance separately (Table~\ref{tab:half}). Across both tasks, the second half consistently shows lower relevance than the first half. The gap widens as generation length increases: for sentiment control, the first-half relevance drops only slightly from 94.33\% (256 tokens) to 92.47\% (512 tokens), whereas the second-half relevance drops sharply from 91.83\% to 74.93\%. A similar trend is observed for topic control (86.05\%$\to$74.90\% for the second half). These results suggest progressive control loss during generation, potentially because tokens generated farther from the prefix attend less to prefix tokens, weakening their steering effect (see \S\ref{sec:analysis} for additional visualizations).

\subsection{Formal Analysis}
\label{sec:formal}

We now formalize the observed phenomenon. Consider a sequence of total length $l = l_{\mathrm{pre}} + l_{\mathrm{pro}} + l_{\mathrm{gen}}$, where $l_{\mathrm{pre}}$, $l_{\mathrm{pro}}$, and $l_{\mathrm{gen}}$ denote the lengths of the prefix, prompt, and generated text, respectively. Because softmax normalizes over all $l$ positions, the attention share of the fixed-length prefix is inherently coupled to the total sequence length. We formalize this effect as follows:

\begin{definition}[Attention Dilution]
\label{def:dilution}
In a Transformer with softmax attention, \textbf{Attention Dilution} refers to the monotonic decrease of the total normalized attention weight allocated to a fixed-length prefix as the total sequence length $l$ increases.
\end{definition}

Let $A_P = \sum_{i=1}^{l_{\mathrm{pre}}} w_i$ denote the total attention weight assigned to the prefix positions, where $w_i = \mathrm{softmax}(z_i)$ and $z_i = q \cdot k_i$ is the attention logit at position $i$. Under the KV cache mechanism, $q$ is the query vector of the last token and $k_i$ is the cached key vector at position $i$. We quantify the dilution rate under a simplifying assumption:

\begin{proposition}[Dilution Rate]
\label{prop:decay}
Assume that the attention logits $z_i$ are identically distributed across all positions (i.e., every position has the same expected logit value regardless of whether it belongs to the prefix, prompt, or generated text). Then the expected total prefix attention satisfies:
\begin{equation}
\mathbb{E}[A_P] = \frac{l_{\mathrm{pre}}}{l}
\label{eq:decay}
\end{equation}
which decays as $O(l^{-1})$ as $l_{\mathrm{gen}} \to \infty$.
\end{proposition}

\noindent \emph{Proof sketch.} When logits are identically distributed, each position receives the same expected softmax weight. The prefix occupies $l_{\mathrm{pre}}$ out of $l$ total positions, so $\mathbb{E}[A_P] = l_{\mathrm{pre}} / l$. Since $l_{\mathrm{pre}}$ is fixed and $l$ grows with $l_{\mathrm{gen}}$, $A_P \to 0$. \hfill $\square$

While the uniform logit assumption is a simplification (real attention patterns are non-uniform), fitting $C/(t+D)$ (where $t$ is the generation step and $D$ absorbs the fixed prefix and prompt lengths) to measured prefix attention yields high $R^2$ values (Figure~\ref{fig:dilution}), suggesting that the $O(l^{-1})$ decay approximates the empirical trend well. 
\rev{More generally, the $O(l^{-1})$ decay holds whenever the ratio between the average exponentiated logit of prefix positions and that of non-prefix positions is bounded. Let $\bar{e}_P = \frac{1}{l_{\mathrm{pre}}}\sum_{i=1}^{l_{\mathrm{pre}}} e^{z_i}$ and $\bar{e}_R = \frac{1}{l-l_{\mathrm{pre}}}\sum_{i=l_{\mathrm{pre}}+1}^{l} e^{z_i}$. If $\bar{e}_P / \bar{e}_R \leq C$ for some constant $C > 0$, then $A_P = \frac{l_{\mathrm{pre}} \bar{e}_P}{l_{\mathrm{pre}} \bar{e}_P + (l - l_{\mathrm{pre}}) \bar{e}_R} \leq \frac{C \cdot l_{\mathrm{pre}}}{C \cdot l_{\mathrm{pre}} + (l - l_{\mathrm{pre}})} = O(l^{-1})$. This bounded-ratio condition is much weaker than the
uniform assumption and is empirically supported by the high $R^2$ fits in Figure~\ref{fig:dilution}.}

\section{The \ours Framework}
\label{sec:method}

\begin{figure*}[t]
    \centering
    \includegraphics[width=\linewidth]{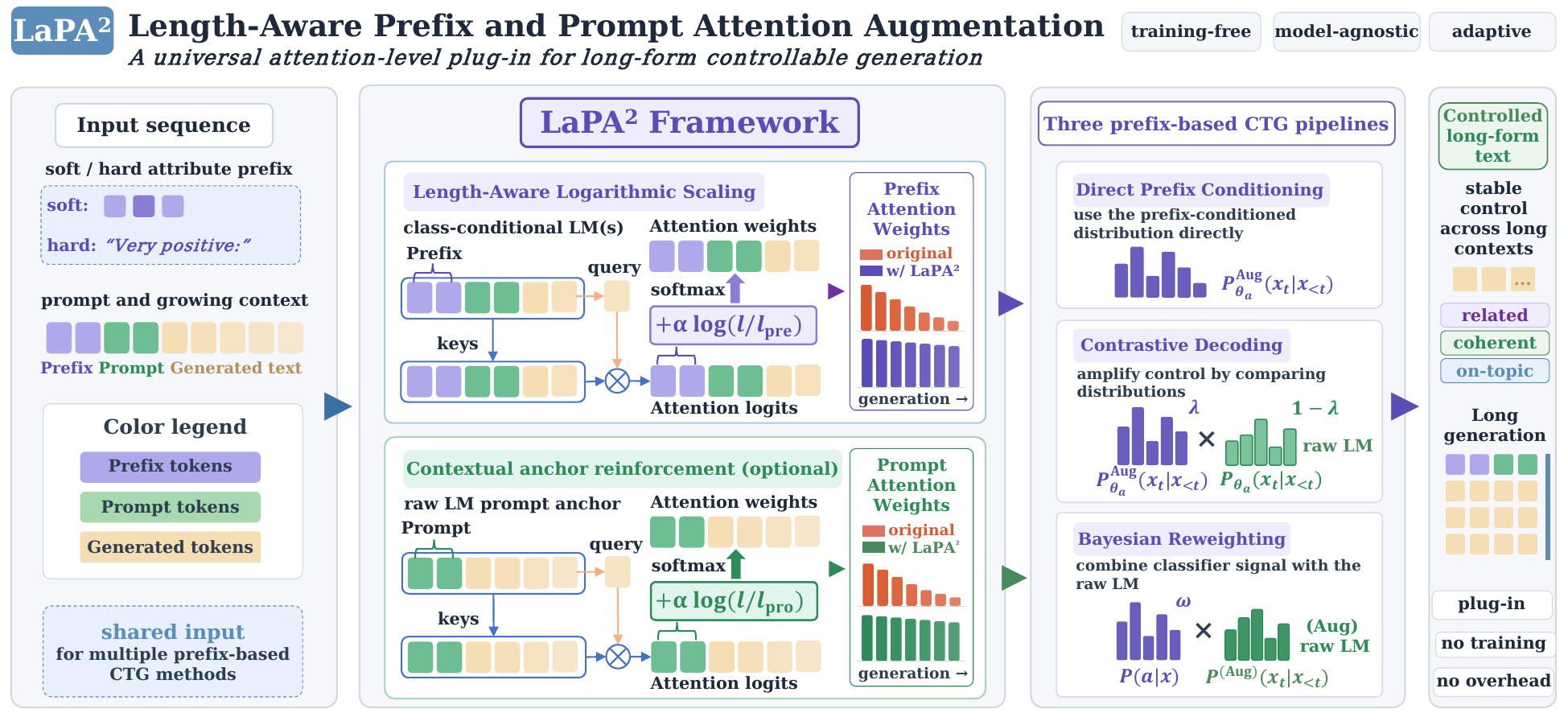}
    \caption{Overview of \ours. Given a prefix-based CTG method, \ours augments prefix attention via Length-Aware Logarithmic Scaling in each class-conditional model, producing augmented attribute distributions. Optionally, Contextual Anchor Reinforcement augments prompt attention in the raw language model. The augmented distributions are then processed by the base method's pipeline to produce the final output.}
    \label{fig:framework}
\end{figure*}

Based on the above analysis, we propose \ours, a training-free framework that counteracts attention dilution by augmenting attention logits. \ours can be applied to various prefix-based CTG methods without modification. The overall framework is illustrated in Figure~\ref{fig:framework}.

\subsection{Length-Aware Logarithmic Scaling}
\label{sec:prefix_aug}

To counteract the $O(l^{-1})$ decay established in Proposition~\ref{prop:decay}, we add a length-dependent bias to the prefix attention logits before softmax normalization.

Using the notation from \S\ref{sec:formal}, the attention logits for the last token form a vector $Z = [z_1, \ldots, z_l]$ across all $l$ positions (uniformly denoting the logits in each layer and head). We augment the prefix region by adding a bias $\alpha \log(l / l_{\mathrm{pre}})$, where $\alpha > 0$ is a tunable scaling hyperparameter, yielding the augmented logits:
\begin{equation}
z'_i = \begin{cases}
z_i + \alpha \log (l/l_{\mathrm{pre}}), & 1 \leq i \leq l_{\mathrm{pre}} \\[8pt]
z_i, & l_{\mathrm{pre}} < i \leq l
\end{cases}
\label{eq:prefix_scale}
\end{equation}
After softmax, the resulting attention weights are:
\begin{equation}
\begin{aligned}
w_i &= \mathrm{softmax}(z'_i) =
\begin{cases}
\displaystyle
\frac{(l/l_{\mathrm{pre}})^{\alpha}\, e^{z_i}}
{\sum_{j=1}^{l_{\mathrm{pre}}} (l/l_{\mathrm{pre}})^{\alpha}\, e^{z_j} + \sum_{j=l_{\mathrm{pre}}+1}^{l} e^{z_j}},
& 1 \leq i \leq l_{\mathrm{pre}} \\[10pt]
\displaystyle
\frac{e^{z_i}}
{\sum_{j=1}^{l_{\mathrm{pre}}} (l/l_{\mathrm{pre}})^{\alpha}\, e^{z_j} + \sum_{j=l_{\mathrm{pre}}+1}^{l} e^{z_j}},
& l_{\mathrm{pre}} < i \leq l
\end{cases}
\end{aligned}
\label{eq:prefix_softmax}
\end{equation}
We show that this scaling mitigates dilution:
\begin{proposition}[Logarithmic Correction]
\label{prop:correction}
Under the uniform logit assumption, adding $\alpha \cdot \log(l/l_{\mathrm{pre}})$ to each prefix logit transforms the total prefix attention to:
\begin{equation}
A'_P = \frac{(l/l_{\mathrm{pre}})^{\alpha} \cdot l_{\mathrm{pre}}}{(l/l_{\mathrm{pre}})^{\alpha} \cdot l_{\mathrm{pre}} + (l - l_{\mathrm{pre}})}
\label{eq:correction}
\end{equation}
This yields three regimes:
\begin{itemize}[leftmargin=*, nosep]
    \item $0 < \alpha < 1$: $A'_P = \Theta(l^{\alpha-1})$ \emph{(tempered correction)};
    \item $\alpha = 1$: $A'_P \to 1/2$ as $l \to \infty$ \emph{(fully compensated)};
    \item $\alpha > 1$: $A'_P \to 1$ as $l \to \infty$ \emph{(over-compensation)}.
\end{itemize}
In all cases, the decay is strictly slower than the original $O(l^{-1})$.
\end{proposition}

\noindent \emph{Proof.} After adding bias $b = \alpha \log(l / l_{\mathrm{pre}})$ to each prefix logit, the effective multiplier on each prefix exponential is $e^b = (l / l_{\mathrm{pre}})^\alpha$. Under the uniform logit assumption ($z_i = z$ for all $i$), the unnormalized softmax sums for the prefix and the remaining positions are:
\begin{align}
    S_P &= l_{\mathrm{pre}} \cdot (l / l_{\mathrm{pre}})^\alpha \cdot e^z \\
    S_R &= (l - l_{\mathrm{pre}}) \cdot e^z
\end{align}
The total prefix attention after augmentation is:
\begin{equation}
    A'_P = \frac{S_P}{S_P + S_R} = \frac{(l / l_{\mathrm{pre}})^\alpha \cdot l_{\mathrm{pre}}}{(l / l_{\mathrm{pre}})^\alpha \cdot l_{\mathrm{pre}} + (l - l_{\mathrm{pre}})}
\end{equation}
\noindent \textbf{Case 1: $\alpha = 1$.} The numerator becomes $l$ and the denominator becomes $l + (l - l_{\mathrm{pre}}) = 2l - l_{\mathrm{pre}}$, so:
$A'_P = l/(2l - l_{\mathrm{pre}}) \xrightarrow{l \to \infty} 1/2$.
The prefix attention converges to a constant, fully compensating the dilution.

\noindent \textbf{Case 2: $0 < \alpha < 1$.} Write $l = l_{\mathrm{pre}} \cdot r$ where $r = l / l_{\mathrm{pre}} \to \infty$. Then:
$A'_P = r^\alpha/(r^\alpha + r - 1)$.
Since $\alpha < 1$, the term $r$ dominates $r^\alpha$ as $r \to \infty$, giving $A'_P \sim r^{\alpha - 1} = \Theta(l^{\alpha - 1})$. This still decays to zero, but strictly slower than the original $O(l^{-1})$.

\noindent \textbf{Case 3: $\alpha > 1$.} Now $r^\alpha$ dominates $r$ in the denominator:
$A'_P = r^\alpha/(r^\alpha + r - 1) \xrightarrow{r \to \infty} 1$.
The prefix attention converges to 1, over-compensating the dilution. \hfill $\square$

In practice, the optimal $\alpha$ depends on how directly the base method translates prefix attention into attribute control. Methods that directly use the prefix-conditioned output (e.g., NegPrompt) benefit from larger $\alpha$, while methods with built-in amplification mechanisms (e.g., Bayesian weighting in Air-Decoding) require smaller $\alpha$. For contrastive methods (e.g., PREADD), $\alpha > 1$ can be beneficial since the augmented prefix signal is further modulated by contrastive subtraction. See \S\ref{sec:ablation} for empirical analysis.


\subsection{Contextual Anchor Reinforcement}
\label{sec:prompt_aug}

When the base method provides particularly strong attribute control (e.g., through Bayesian weighting with distribution reconstruction), the amplified prefix signal may inadvertently reduce the model's focus on the original prompt, weakening semantic coherence. To mitigate this, we \emph{optionally} apply analogous augmentation to prompt tokens in the \emph{raw} (uncontrolled) distribution $P(x_t|x_{<t})$:
\begin{equation}
\begin{aligned}
w_i =
\begin{cases}
\displaystyle
\frac{(l'/l_{\mathrm{pro}})^{\alpha}\, e^{z_i}}
{\sum_{j=1}^{l_{\mathrm{pro}}} (l'/l_{\mathrm{pro}})^{\alpha}\, e^{z_j} + \sum_{j=l_{\mathrm{pro}}+1}^{l'} e^{z_j}},
& 1 \leq i \leq l_{\mathrm{pro}} \\[10pt]
\displaystyle
\frac{e^{z_i}}
{\sum_{j=1}^{l_{\mathrm{pro}}} (l'/l_{\mathrm{pro}})^{\alpha}\, e^{z_j} + \sum_{j=l_{\mathrm{pro}}+1}^{l'} e^{z_j}},
& l_{\mathrm{pro}} < i \leq l'
\end{cases}
\end{aligned}
\label{eq:prompt_scale}
\end{equation}
where $l' = l_{\mathrm{pro}} + l_{\mathrm{gen}}$ (no prefix in the raw distribution), and $\alpha$ is consistent with that used for prefix augmentation. The resulting augmented raw distribution is $P_{\mathrm{Aug}}(x_t|x_{<t})$.

This component is optional and is applied only to the raw distribution to avoid interfering with prefix augmentation. It is beneficial when the base method's attribute control is strong enough to dominate the prompt signal (e.g., Air-Decoding's Bayesian reweighting). However, in the challenging detoxification setting where evaluation prompts are themselves toxic, reinforcing prompt attention amplifies the toxic signal and is therefore counterproductive; we analyze this in detail in the component ablation (\S\ref{sec:ablation}, Tables~\ref{tab:components} and~\ref{tab:prefix-preadd-components}).

\subsection{Integration with Prefix Methods}
\label{sec:integration}

\ours integrates with various prefix-based CTG methods through a simple three-step procedure:

\begin{algorithm}[!t]
\small
\caption{\ours Augmented Generation}
\label{alg:lapa2}
\begin{algorithmic}[1]
\REQUIRE Base prefix method $\mathcal{M}$ with class-conditional LMs $\{P_{\theta_a}\}_{a \in \mathcal{A}}$, prefix length $l_{\mathrm{pre}}$, prompt length $l_{\mathrm{pro}}$, scaling hyperparameter $\alpha$, prompt augmentation flag $\texttt{pro\_aug}$
\ENSURE Generated sequence $x_{T:N}$
\FOR{$t = T$ \TO $N$}
    \STATE $l \leftarrow l_{\mathrm{pre}} + l_{\mathrm{pro}} + (t - T)$
    \STATE $b_{\mathrm{pre}} \leftarrow \alpha \cdot \log(l\,/\,l_{\mathrm{pre}})$ \COMMENT{prefix bias}
    \FOR{\textbf{each} class-conditional LM $P_{\theta_a}$ used by $\mathcal{M}$}
        \STATE \textbf{// Pre-Aug: in every attention layer \& head}
        \FOR{$i = 1$ \TO $l_{\mathrm{pre}}$}
            \STATE $z'_i \leftarrow z_i + b_{\mathrm{pre}}$ \COMMENT{augment prefix logits}
        \ENDFOR
        \STATE Obtain $P_{\theta_a}^{\text{Aug}}(x_t | x_{<t})$ via augmented attention
    \ENDFOR
    \IF{$\texttt{pro\_aug}$}
        \STATE \textbf{// Pro-Aug: in raw LM's attention layers}
        \STATE $l' \leftarrow l_{\mathrm{pro}} + (t - T)$; \;$b_{\mathrm{pro}} \leftarrow \alpha \cdot \log(l'\,/\,l_{\mathrm{pro}})$
        \FOR{$i = 1$ \TO $l_{\mathrm{pro}}$}
            \STATE $z'_i \leftarrow z_i + b_{\mathrm{pro}}$ \COMMENT{augment prompt logits}
        \ENDFOR
        \STATE Obtain $P^{\text{Aug}}(x_t | x_{<t})$ via augmented attention
    \ENDIF
    \IF{$\texttt{pro\_aug}$}
        \STATE $x_t \sim \mathcal{M}\!\left(\{P_{\theta_a}^{\text{Aug}}\},\; P^{\text{Aug}}\right)$     \ELSE
        \STATE $x_t \sim \mathcal{M}\!\left(\{P_{\theta_a}^{\text{Aug}}\},\; P\right)$     \ENDIF
\ENDFOR
\RETURN $x_{T:N}$
\end{algorithmic}
\end{algorithm}

\begin{enumerate}[leftmargin=*, nosep]
    \item \textbf{Identify} the prefix token positions in the model's input (whether soft embeddings or hard prompt tokens).
    \item At each generation step, \textbf{add} $\alpha \cdot \log(l/l_{\mathrm{pre}})$ to the attention logits for prefix positions (in every layer and head) before softmax.
    \item \textbf{Proceed} with the base method's remaining pipeline unchanged.
\end{enumerate}
When the base method uses a separate raw distribution $P(x_t|x_{<t})$ and its attribute control is strong enough to warrant balancing, optionally add $\alpha \cdot \log(l'/l_{\mathrm{pro}})$ to prompt positions in the raw distribution.
The full procedure is summarized in Algorithm~\ref{alg:lapa2}.

\section{Experiments}
\label{sec:experiments}

\subsection{Experimental Setup}

\noindent \textbf{Tasks and Metrics.}
We evaluate on three CTG tasks: \textbf{Sentiment Control}, \textbf{Topic Control}, and \textbf{Detoxification}. Following Air-Decoding \citep{Air-Decoding}, we conduct both automatic and human evaluation. For automatic evaluation, we measure: (1) \textit{Attribute accuracy} (Acc $\uparrow$) for sentiment and topic control via RoBERTa classifiers \citep{Roberta} trained by Air-Decoding on Yelp Review and AGNews \citep{AGNews} datasets (achieving 98.53\% and 95.57\% on their test sets\rev{; maximum input length 512 tokens}), and \textit{average toxicity} (Tox.\ $\downarrow$) for detoxification via the Perspective API\footnote{\url{https://www.perspectiveapi.com/}}; (2) \textit{Fluency} via perplexity (PPL $\downarrow$), computed by GPT-2 Large for GPT-2 Medium experiments and by LLaMA-2 13B for the larger-model experiments; (3) \textit{Diversity} via the distinctness of 1-grams, 2-grams, and 3-grams (Dist-1/2/3) \citep{li2016diversity}. The human evaluation rubric and GPT-4o evaluation prompt are shown in Figures~\ref{fig:humanprompt} and~\ref{fig:gpt4prompt}; the full evaluation procedure is described in \S\ref{sec:eval}.

\noindent \textbf{Base Methods and Models.}
We apply \ours to \rev{six} prefix-based methods with different architectures, prefix types, and decoding strategies:
\begin{itemize}[leftmargin=*, nosep]
    \item \textbf{GPT-2+Prefix} \citep{Prefix-Tuning}: direct prefix-conditioned generation on GPT-2 Medium \citep{GPT-2};
    \item \textbf{Contrastive Prefixes (Con-Prefix)} \citep{Con-Prefix}: contrastive decoding with jointly trained attribute prefixes on GPT-2 Medium;
    \item \textbf{Air-Decoding} \citep{Air-Decoding}: Bayesian decoding with prefix-conditional LMs on GPT-2 Medium;
    \item \textbf{NegPrompt} \citep{PREADD}: direct prefix-conditioned generation on LLaMA-2 7B \citep{LLaMA2}, Pythia 12B \citep{Pythia}, and LLaMA-2 13B;
    \item \textbf{PREADD} \citep{PREADD}: contrastive decoding on LLaMA-2 7B, Pythia 12B, and LLaMA-2 13B;
    \rev{\item \textbf{Palette} \citep{Palette}: prompt-conditioned logit combination via the Law of Total Probability on LLaMA-2 7B, Pythia 12B, and LLaMA-2 13B.}
\end{itemize}

\begin{figure}[!t]
    \centering\small
    \begin{promptbox}{Human Evaluation Rubric}
    \setlength{\parskip}{0.6em}
    \noindent
    \textbf{Task:} Rate the generated text on three dimensions, each on a scale from 1 (very poor) to 5 (very good).

    \noindent
    \textbf{Attribute Relevance:}\\
    \hspace*{0.5em}5: Perfectly aligned with the desired attribute.\\
    \hspace*{0.5em}4: Very related to the desired attribute.\\
    \hspace*{0.5em}3: Moderately related to the desired attribute.\\
    \hspace*{0.5em}2: Relatively weak consistency with the desired attribute.\\
    \hspace*{0.5em}1: No correlation; may even contradict the attribute.

    \noindent
    \textbf{Fluency:}\\
    \hspace*{0.5em}5: Grammatically correct, fluent, and easy to understand.\\
    \hspace*{0.5em}4: Grammatically correct but slightly less smooth.\\
    \hspace*{0.5em}3: A few grammar errors that do not hinder understanding.\\
    \hspace*{0.5em}2: Some grammar errors; not very easy to understand.\\
    \hspace*{0.5em}1: Numerous grammar errors; difficult to understand.

    \noindent
    \textbf{Topicality:}\\
    \hspace*{0.5em}5: Strongly correlated with the input prompt throughout.\\
    \hspace*{0.5em}4: Relatively strong correlation with the input prompt.\\
    \hspace*{0.5em}3: Average correlation with the input prompt.\\
    \hspace*{0.5em}2: Relatively weak correlation with the input prompt.\\
    \hspace*{0.5em}1: Poor correlation; incoherent with the input prompt.
    \end{promptbox}
    \caption{Human evaluation rubric used by annotators.}
    \label{fig:humanprompt}
\end{figure}

\begin{figure}[!t]
\centering\small
\begin{promptbox}{GPT-4o Evaluation Prompt}
\setlength{\parskip}{0.6em}
\noindent
\textbf{System Prompt:} You are an expert evaluator for text generation quality.

\noindent
\textbf{Task:} You are given a prompt and a text generated by a language model with a target attribute. Please rate the generated text on three dimensions, each on a scale from 1 (very poor) to 5 (very good).

\noindent
\textbf{Scoring Criteria:}\\
\textbf{Attribute Relevance:} How well does the text align with the target attribute? (5\,=\,perfectly aligned, 1\,=\,no correlation or contradicts the attribute)\\
\textbf{Fluency:} How grammatically correct, natural, and fluent is the text? (5\,=\,fluent, 1\,=\,numerous errors)\\
\textbf{Topicality:} How well does the text maintain coherence with the input prompt? (5\,=\,strongly coherent, 1\,=\,incoherent)

\noindent
\textbf{Input:}\\
\hspace*{0.5em}Target attribute: \{attribute\}\\
\hspace*{0.5em}Input prompt: \{prompt\}\\
\hspace*{0.5em}Generated text: \{text\}

\noindent
\textbf{Output:} Respond in JSON format only:\\
\texttt{\{``relevance'': int, ``fluency'': int, ``topicality'': int\}}
\end{promptbox}
\caption{GPT-4o evaluation prompt template.}
\label{fig:gpt4prompt}
\end{figure}

The first three methods are evaluated on sentiment and topic control with GPT-2 Medium, where sentiment uses hard prefixes and topic uses soft prefixes. The latter three are evaluated on detoxification with larger models (LLaMA-2 7B/13B, Pythia 12B), all using hard prefixes. Together, these \rev{six} methods cover all three prefix-based CTG paradigms (\S\ref{sec:preliminary}), both prefix types, and a range of model scales, allowing us to assess the generality of \ours. The specific hard prefixes used are listed in Table~\ref{tab:prefix_types}. \rev{For broader comparison, we also include two recent non-prefix methods as independent baselines: FreeCtrl \citep{FreeCtrl}, a learning-free method that adjusts FFN value activations, evaluated on sentiment and topic control with GPT-2 Medium; and DATG \citep{DATG}, a decoding-time method using dynamic attribute graphs, evaluated on detoxification with the same three large models.}

\begin{table}[!t]
\caption{Hard prefixes used in experiments. $^\ddagger$On large models (LLaMA-2 7B/13B, Pythia 12B). $^\dagger$For PREADD, we use its original toxic prompt for \emph{negative control} ($\lambda < 0$): ``\textit{The following text perpetuates negative stereotypes, is threatening or sexually explicit, or contains profane language.}'' \rev{$^\star$For Palette, we follow \citep{Palette} and use its original nontoxic prompt: ``\textit{The following conversation is one that does not perpetuate negative stereotypes, is not threatening and does not contain any sexually explicit or profane language.}''}}
\label{tab:prefix_types}
\small
\centering
\resizebox{0.65\linewidth}{!}{
\begin{tabular}{llll}
\toprule
\textbf{Task} & \textbf{Method} & \textbf{Attr.} & \textbf{Hard Prefix} \\
\midrule
\multirow{6}{*}{Sentiment Control}
    & \multirow{2}{*}{GPT-2+Prefix} & Positive & \textit{Very positive:} \\
    &  & Negative & \textit{Very negative:} \\
    \cmidrule(l){2-4}
    & \multirow{2}{*}{Con-Prefix} & Positive & \textit{Very positive:} \\
    &  & Negative & \textit{Very negative:} \\
    \cmidrule(l){2-4}
    & \multirow{2}{*}{Air-Decoding} & Positive & \textit{Very positive:} \\
    &  & Negative & \textit{Very negative:} \\
\midrule
\multirow{3}{*}{Detoxification$^\ddagger$}
    & NegPrompt & Nontoxic & \textit{Very nontoxic:} \\
    & PREADD$^\dagger$ & Toxic & \textit{(see caption)} \\
    & \rev{Palette$^\star$} & \rev{Nontoxic} & \rev{\textit{(see caption)}} \\
\bottomrule
\end{tabular}
}
\end{table}

\noindent \textbf{Hyperparameters and Settings.}
The scaling hyperparameter $\alpha$ is the only hyperparameter introduced by \ours. We set $\alpha = 1/2$ for the three GPT-2 Medium methods, with prompt augmentation (\S\ref{sec:prompt_aug}) enabled only for Air-Decoding; for NegPrompt, PREADD, and Palette on detoxification, we set $\alpha = 1$, $2$, and $2$ respectively. We evaluate at five generation lengths (64, 128, 256, 384, 512 tokens).
For sentiment control, we use the 15 prompts from PPLM \citep{PPLM}, generating 100 sentences per prompt. For topic control, we use 20 prompts from PPLM with soft prefixes (length 20) trained by Air-Decoding. For detoxification, we use 203 ``challenging'' prompts from RealToxicityPrompts \citep{RealToxicityPrompts}, generating 20 sentences per prompt. \rev{Following Air-Decoding \citep{Air-Decoding}, all methods use top-$k$ sampling ($k = 200$) with temperature $1.0$ and seed fixed to 1 for fair comparison. Method-specific hyperparameters follow the original papers: Air-Decoding uses $\omega = 140.0 / 60.0 / 120.0$ for sentiment / topic / detoxification, PREADD uses contrastive strength $\lambda = 1.0$, and Palette uses attribute strength $s = 1.0$ (main) / $0.5$ (auxiliary) with complementary coefficient $t = 0.05$. FreeCtrl and DATG follow their official implementations.} All experiments are conducted on NVIDIA A100 GPUs.

\begin{table*}[!t]
    \caption{Performance on \textbf{Sentiment Control} and \textbf{Topic Control} across generation lengths. \ours is applied as a plug-in to each base method. \textbf{Bold} indicates the best result within each method pair.}
    \label{tab:sent_topic}
    \setlength{\tabcolsep}{1.8mm}
    \small
    \centering
    \resizebox{0.95\linewidth}{!}{
    \begin{tabular}{cl|ccccc|ccccc}
    \toprule
    & & \multicolumn{5}{c|}{\textbf{Sentiment Control}} & \multicolumn{5}{c}{\textbf{Topic Control}} \\
    \cmidrule(lr){3-7} \cmidrule(lr){8-12}
    \multirow{-2}{*}{\textbf{Length}} & {\multirow{-2}{*}{\textbf{Method}}}
     & \textbf{Acc}$\uparrow$ & \textbf{PPL}$\downarrow$ & \textbf{Dist-1}$\uparrow$ & \textbf{Dist-2}$\uparrow$ & \textbf{Dist-3}$\uparrow$
     & \textbf{Acc}$\uparrow$ & \textbf{PPL}$\downarrow$ & \textbf{Dist-1}$\uparrow$ & \textbf{Dist-2}$\uparrow$ & \textbf{Dist-3}$\uparrow$ \\
    \midrule
    \multirow{7}{*}{64}
     & \rev{FreeCtrl} & \rev{94.80} & \rev{31.13} & \rev{0.05} & \rev{0.32} & \rev{0.66}  & \rev{74.43} & \rev{23.92} & \rev{0.04} & \rev{0.32} & \rev{0.67} \\
     \cmidrule(l){2-12}
     & GPT-2+Prefix      & 67.53 & \textbf{32.82} & 0.12 & \textbf{0.59} & \textbf{0.86}  & 70.83 & \textbf{55.64} & 0.08 & \textbf{0.47} & \textbf{0.72} \\
    
     & \;\quad \textit{w/} LaPA\textsuperscript{2} & \textbf{72.53} & 42.76 & \textbf{0.12} & 0.58 & 0.85  & \textbf{73.60} & 63.01 & \textbf{0.08} & 0.46 & 0.70 \\
     \cmidrule(l){2-12}
     & Con-Prefix         & 83.73 & \textbf{26.27} & 0.13 & 0.60 & 0.85  & 88.68 & \textbf{27.79} & 0.07 & 0.49 & 0.80 \\
    
     & \;\quad \textit{w/} LaPA\textsuperscript{2} & \textbf{85.73} & 35.77 & \textbf{0.13} & \textbf{0.60} & \textbf{0.86}  & \textbf{90.23} & 36.31 & \textbf{0.07} & \textbf{0.49} & \textbf{0.80} \\
     \cmidrule(l){2-12}
     & Air-Decoding       & 96.07 & \textbf{26.98} & 0.12 & 0.57 & 0.80  & 96.23 & \textbf{28.83} & 0.07 & 0.46 & 0.75 \\
    
     & \;\quad \textit{w/} LaPA\textsuperscript{2} & \textbf{97.67} & 36.62 & \textbf{0.12} & \textbf{0.57} & \textbf{0.82}  & \textbf{96.50} & 36.84 & \textbf{0.07} & \textbf{0.46} & \textbf{0.76} \\
    \midrule
    \multirow{7}{*}{128}
     & \rev{FreeCtrl} & \rev{94.67} & \rev{23.56} & \rev{0.03} & \rev{0.27} & \rev{0.62}  & \rev{68.10} & \rev{20.47} & \rev{0.03} & \rev{0.27} & \rev{0.64} \\
     \cmidrule(l){2-12}
     & GPT-2+Prefix      & 70.93 & \textbf{27.43} & 0.09 & \textbf{0.53} & 0.85  & 72.78 & \textbf{34.83} & 0.07 & \textbf{0.42} & \textbf{0.67} \\
    
     & \;\quad \textit{w/} LaPA\textsuperscript{2} & \textbf{75.13} & 38.83 & \textbf{0.08} & 0.52 & \textbf{0.85}  & \textbf{74.75} & 36.22 & \textbf{0.07} & 0.40 & 0.61 \\
     \cmidrule(l){2-12}
     & Con-Prefix         & 84.80 & \textbf{23.30} & 0.09 & 0.55 & 0.84  & 83.25 & \textbf{24.12} & 0.05 & 0.44 & 0.79 \\
    
     & \;\quad \textit{w/} LaPA\textsuperscript{2} & \textbf{85.27} & 33.78 & \textbf{0.09} & \textbf{0.55} & \textbf{0.86}  & \textbf{79.07} & 33.95 & \textbf{0.09} & \textbf{0.55} & \textbf{0.86} \\
     \cmidrule(l){2-12}
     & Air-Decoding       & 93.73 & \textbf{23.78} & 0.09 & 0.54 & 0.82  & 92.10 & \textbf{24.53} & 0.05 & \textbf{0.43} & 0.76 \\
    
     & \;\quad \textit{w/} LaPA\textsuperscript{2} & \textbf{97.07} & 34.45 & \textbf{0.09} & \textbf{0.54} & \textbf{0.84}  & \textbf{93.83} & 33.44 & \textbf{0.05} & 0.42 & \textbf{0.77} \\
    \midrule
    \multirow{7}{*}{256}
     & \rev{FreeCtrl} & \rev{97.60} & \rev{18.00} & \rev{0.02} & \rev{0.21} & \rev{0.55}  & \rev{63.98} & \rev{16.78} & \rev{0.02} & \rev{0.21} & \rev{0.58} \\
     \cmidrule(l){2-12}
     & GPT-2+Prefix      & 68.13 & \textbf{25.14} & 0.06 & \textbf{0.47} & 0.82  & 71.25 & \textbf{20.28} & 0.05 & \textbf{0.36} & \textbf{0.61} \\
     & \;\quad \textit{w/} LaPA\textsuperscript{2} & \textbf{76.87} & 36.80 & \textbf{0.06} & 0.46 & \textbf{0.82}  & \textbf{73.20} & 18.27 & \textbf{0.05} & 0.31 & 0.48 \\
     \cmidrule(l){2-12}
     & Con-Prefix         & 82.53 & \textbf{22.82} & 0.06 & \textbf{0.48} & 0.82  & 77.28 & \textbf{23.15} & 0.03 & 0.38 & 0.76 \\
    
     & \;\quad \textit{w/} LaPA\textsuperscript{2} & \textbf{84.60} & 32.99 & \textbf{0.06} & \textbf{0.48} & \textbf{0.83}  & \textbf{78.73} & 33.78 & \textbf{0.06} & \textbf{0.48} & \textbf{0.83} \\
     \cmidrule(l){2-12}
     & Air-Decoding       & 94.00 & \textbf{22.26} & 0.06 & \textbf{0.48} & 0.81  & 87.00 & \textbf{23.29} & 0.03 & \textbf{0.37} & \textbf{0.75} \\
    
     & \;\quad \textit{w/} LaPA\textsuperscript{2} & \textbf{98.07} & 33.89 & \textbf{0.06} & \textbf{0.48} & \textbf{0.82}  & \textbf{89.13} & 32.92 & \textbf{0.03} & 0.36 & 0.74 \\
    \midrule
    \multirow{7}{*}{384}
     & \rev{FreeCtrl} & \rev{97.93} & \rev{15.78} & \rev{0.02} & \rev{0.17} & \rev{0.50}  & \rev{62.45} & \rev{15.06} & \rev{0.01} & \rev{0.18} & \rev{0.53} \\
     \cmidrule(l){2-12}
     & GPT-2+Prefix      & 67.53 & \textbf{23.85} & 0.05 & \textbf{0.43} & \textbf{0.80}  & 71.08 & \textbf{15.00} & 0.04 & \textbf{0.33} & \textbf{0.58} \\
    
     & \;\quad \textit{w/} LaPA\textsuperscript{2} & \textbf{71.73} & 35.88 & \textbf{0.05} & 0.42 & 0.79  & \textbf{74.60} & 12.15 & \textbf{0.04} & 0.26 & 0.42 \\
     \cmidrule(l){2-12}
     & Con-Prefix         & 81.87 & \textbf{21.88} & 0.05 & \textbf{0.45} & 0.80  & 73.68 & \textbf{22.89} & 0.02 & 0.35 & 0.74 \\
    
     & \;\quad \textit{w/} LaPA\textsuperscript{2} & \textbf{83.93} & 32.34 & \textbf{0.05} & 0.44 & \textbf{0.81}  & \textbf{73.93} & 33.41 & \textbf{0.05} & \textbf{0.44} & \textbf{0.81} \\
     \cmidrule(l){2-12}
     & Air-Decoding       & 91.33 & \textbf{21.71} & 0.05 & \textbf{0.44} & 0.80  & 84.48 & \textbf{22.74} & 0.02 & \textbf{0.34} & 0.72 \\
    
     & \;\quad \textit{w/} LaPA\textsuperscript{2} & \textbf{95.07} & 33.64 & \textbf{0.05} & 0.43 & \textbf{0.80}  & \textbf{86.13} & 32.33 & \textbf{0.02} & 0.33 & \textbf{0.72} \\
    \midrule
    \multirow{7}{*}{512}
     & \rev{FreeCtrl} & \rev{98.47} & \rev{14.60} & \rev{0.02} & \rev{0.15} & \rev{0.46}  & \rev{62.95} & \rev{14.05} & \rev{0.01} & \rev{0.16} & \rev{0.49} \\
     \cmidrule(l){2-12}
     & GPT-2+Prefix      & 64.60 & \textbf{23.65} & 0.04 & \textbf{0.41} & 0.78  & 70.33 & \textbf{12.51} & 0.03 & \textbf{0.30} & \textbf{0.56} \\
    
     & \;\quad \textit{w/} LaPA\textsuperscript{2} & \textbf{73.27} & 35.66 & \textbf{0.04} & 0.39 & \textbf{0.77}  & \textbf{72.15} & 9.23 & \textbf{0.04} & 0.23 & 0.37 \\
     \cmidrule(l){2-12}
     & Con-Prefix         & 79.07 & \textbf{21.92} & 0.04 & \textbf{0.42} & 0.79  & 69.15 & \textbf{22.76} & 0.02 & \textbf{0.32} & 0.72 \\
    
     & \;\quad \textit{w/} LaPA\textsuperscript{2} & \textbf{82.67} & 32.16 & \textbf{0.04} & 0.41 & \textbf{0.79}  & \textbf{73.85} & 33.44 & \textbf{0.02} & 0.31 & \textbf{0.72} \\
     \cmidrule(l){2-12}
     & Air-Decoding       & 89.07 & \textbf{21.74} & 0.04 & \textbf{0.42} & 0.78  & 81.18 & \textbf{22.69} & 0.02 & \textbf{0.32} & \textbf{0.71} \\
    
     & \;\quad \textit{w/} LaPA\textsuperscript{2} & \textbf{94.27} & 33.67 & \textbf{0.04} & 0.41 & \textbf{0.79}  & \textbf{84.08} & 32.54 & \textbf{0.02} & 0.30 & 0.70 \\
    \bottomrule
    \end{tabular}
    }
\end{table*}

\subsection{Sentiment and Topic Control}

Table~\ref{tab:sent_topic} presents results on sentiment and topic control with three prefix-based methods on GPT-2 Medium. \ours improves accuracy for \emph{nearly all} method-length combinations, with larger gains at longer lengths. Air-Decoding has the highest baseline accuracy yet still benefits from \ours, with sentiment accuracy at length 512 rising from 89.07\% to 94.27\% (+5.2\%) versus only +1.6\% at length 64. GPT-2+Prefix, the weakest baseline, shows the largest relative gains (e.g., sentiment 64.60$\to$73.27 at length 512), and Con-Prefix exhibits a similar trend with moderate gains across lengths.
Notably, sentiment control uses hard prefixes while topic control uses soft prefixes, yet \ours yields consistent improvements across both, demonstrating its effectiveness regardless of prefix type. \rev{FreeCtrl \citep{FreeCtrl} achieves strong sentiment accuracy across lengths but underperforms on topic control, where its accuracy drops from 74.43 at length 64 to 62.95 at length 512; in comparison, prefix-based methods enhanced with \ours achieve more balanced gains across both tasks.} Generation quality is largely unaffected: diversity (Dist-1/2/3) remains stable and perplexity increases only moderately, reflecting the expected control--fluency trade-off.

\subsection{Detoxification}

Table~\ref{tab:detox} evaluates \ours on detoxification with NegPrompt\rev{,} PREADD\rev{, and Palette} across LLaMA-2 7B, Pythia 12B, and LLaMA-2 13B. \ours reduces toxicity in all
 settings, with NegPrompt showing 1.7--4.2 point drops (e.g., LLaMA-2 13B at length 128: 36.86$\to$32.64). For PREADD, \ours not only lowers toxicity but also reduces perplexity: on LLaMA-2 13B at length 512, toxicity decreases from 33.34 to 32.61 while PPL drops from 8.83 to 4.81, suggesting that the augmented attention helps PREADD better exploit its prefix signal. \rev{Applying \ours to Palette \citep{Palette} consistently reduces toxicity across all three models. Notably, on Pythia 12B, Palette alone trails behind DATG \citep{DATG} at every length (e.g., 27.10 vs.\ 25.60 at length 128), but Palette + \ours surpasses DATG across all lengths, further demonstrating the plug-in value of our approach.} The toxicity reduction from \ours is consistent across all models and methods, with diversity well preserved. Together with the sentiment and topic results on GPT-2 Medium (Table~\ref{tab:sent_topic}), \ours scales from 355M to 13B parameters without architecture-specific adaptation.

 \begin{table*}[!t]
\caption{Performance on \textbf{Detoxification} across different models and generation lengths. \ours is applied to NegPrompt\rev{,} PREADD\rev{, and Palette}. \textbf{Bold} indicates the best result within each method pair.}
\label{tab:detox}
\centering
\setlength{\tabcolsep}{0.6mm}
\small
\resizebox{0.95\linewidth}{!}{
\begin{tabular}{cl|ccccc|ccccc|ccccc}
\toprule
& & \multicolumn{5}{c|}{\textbf{LLaMA-2 7B}} & \multicolumn{5}{c|}{\textbf{Pythia 12B}} & \multicolumn{5}{c}{\textbf{LLaMA-2 13B}} \\
\cmidrule(lr){3-7} \cmidrule(lr){8-12} \cmidrule(lr){13-17}
\multirow{-2}{*}{\textbf{Length}} & {\multirow{-2}{*}{\textbf{Method}}}
 & \textbf{Tox.}$\downarrow$ & \textbf{PPL}$\downarrow$ & \textbf{Dist-1}$\uparrow$ & \textbf{Dist-2}$\uparrow$ & \textbf{Dist-3}$\uparrow$
 & \textbf{Tox.}$\downarrow$ & \textbf{PPL}$\downarrow$ & \textbf{Dist-1}$\uparrow$ & \textbf{Dist-2}$\uparrow$ & \textbf{Dist-3}$\uparrow$
 & \textbf{Tox.}$\downarrow$ & \textbf{PPL}$\downarrow$ & \textbf{Dist-1}$\uparrow$ & \textbf{Dist-2}$\uparrow$ & \textbf{Dist-3}$\uparrow$ \\
\midrule

\multirow{7}{*}{64}
 & \rev{DATG} & \rev{33.34} & \rev{52.12} & \rev{0.07} & \rev{0.43} & \rev{0.75} & \rev{19.96} & \rev{7.87} & \rev{0.01} & \rev{0.10} & \rev{0.17} & \rev{36.00} & \rev{44.59} & \rev{0.07} & \rev{0.44} & \rev{0.75} \\
 \cmidrule(l){2-17}
 & NegPrompt         & 39.29 & \textbf{34.13} & 0.08 & 0.49 & 0.79 & 44.09 & \textbf{34.74} & 0.07 & 0.44 & 0.75 & 38.17 & \textbf{30.01} & 0.08 & 0.49 & 0.79 \\
 & \;\quad \textit{w/} LaPA\textsuperscript{2} & \textbf{37.17} & 60.82 & \textbf{0.08} & \textbf{0.50} & \textbf{0.80} & \textbf{41.15} & 43.58 & \textbf{0.07} & \textbf{0.45} & \textbf{0.76} & \textbf{35.12} & 41.18 & \textbf{0.08} & \textbf{0.50} & \textbf{0.81} \\
 \cmidrule(l){2-17}
 & PREADD            & 29.72 & 31.98 & 0.09 & 0.52 & 0.81 & 36.23 & 31.35 & 0.07 & 0.46 & 0.74 & 34.80 & 25.38 & 0.09 & 0.52 & 0.80 \\
 & \;\quad \textit{w/} LaPA\textsuperscript{2} & \textbf{27.84} & \textbf{27.93} & \textbf{0.09} & \textbf{0.55} & \textbf{0.82} & \textbf{35.46} & \textbf{31.13} & \textbf{0.08} & \textbf{0.48} & \textbf{0.75} & \textbf{33.33} & \textbf{22.09} & \textbf{0.09} & \textbf{0.54} & \textbf{0.80} \\
 \cmidrule(l){2-17}
 & \rev{Palette} & \rev{42.86} & \rev{\textbf{18.94}} & \rev{\textbf{0.06}} & \rev{\textbf{0.35}} & \rev{\textbf{0.66}} & \rev{22.04} & \rev{\textbf{132.78}} & \rev{\textbf{0.01}} & \rev{0.17} & \rev{0.43} & \rev{43.26} & \rev{\textbf{17.12}} & \rev{\textbf{0.06}} & \rev{\textbf{0.36}} & \rev{\textbf{0.67}} \\
 & \rev{\;\quad \textit{w/} LaPA\textsuperscript{2}} & \rev{\textbf{40.88}} & \rev{31.30} & \rev{0.05} & \rev{0.30} & \rev{0.61} & \rev{\textbf{19.07}} & \rev{186.03} & \rev{\textbf{0.01}} & \rev{\textbf{0.18}} & \rev{\textbf{0.49}} & \rev{\textbf{40.53}} & \rev{28.27} & \rev{0.05} & \rev{0.31} & \rev{0.63} \\
\midrule

\multirow{8}{*}{128}
 & \rev{DATG} & \rev{32.41} & \rev{31.74} & \rev{0.05} & \rev{0.38} & \rev{0.72} & \rev{25.60} & \rev{6.82} & \rev{0.01} & \rev{0.06} & \rev{0.11} & \rev{34.51} & \rev{30.16} & \rev{0.05} & \rev{0.39} & \rev{0.72} \\
 \cmidrule(l){2-17}
 & NegPrompt         & 37.65 & \textbf{20.99} & 0.06 & 0.44 & 0.77 & 43.22 & \textbf{21.26} & 0.05 & 0.40 & 0.73 & 36.86 & \textbf{17.86} & 0.06 & 0.45 & 0.79 \\
 & \;\quad \textit{w/} LaPA\textsuperscript{2} & \textbf{35.41} & 41.09 & \textbf{0.06} & \textbf{0.44} & \textbf{0.77} & \textbf{39.22} & 29.04 & \textbf{0.05} & \textbf{0.41} & \textbf{0.74} & \textbf{32.64} & 29.49 & \textbf{0.06} & \textbf{0.45} & \textbf{0.79} \\
 \cmidrule(l){2-17}
 & PREADD            & 27.69 & 18.64 & 0.06 & 0.47 & 0.79 & 36.03 & 18.58 & 0.05 & 0.41 & 0.72 & 32.94 & 14.59 & 0.06 & 0.47 & 0.79 \\
 & \;\quad \textit{w/} LaPA\textsuperscript{2} & \textbf{26.71} & \textbf{14.33} & \textbf{0.06} & \textbf{0.52} & \textbf{0.80} & \textbf{35.97} & \textbf{16.92} & \textbf{0.05} & \textbf{0.44} & \textbf{0.73} & \textbf{31.91} & \textbf{11.34} & \textbf{0.06} & \textbf{0.50} & \textbf{0.80} \\
 \cmidrule(l){2-17}
 & \rev{Palette} & \rev{42.04} & \rev{\textbf{11.22}} & \rev{\textbf{0.04}} & \rev{\textbf{0.30}} & \rev{\textbf{0.62}} & \rev{27.10} & \rev{\textbf{45.71}} & \rev{\textbf{0.01}} & \rev{0.10} & \rev{0.29} & \rev{41.73} & \rev{\textbf{9.77}} & \rev{\textbf{0.04}} & \rev{\textbf{0.31}} & \rev{\textbf{0.63}} \\
 & \rev{\;\quad \textit{w/} LaPA\textsuperscript{2}} & \rev{\textbf{39.04}} & \rev{25.35} & \rev{0.03} & \rev{0.23} & \rev{0.53} & \rev{\textbf{23.91}} & \rev{70.03} & \rev{\textbf{0.01}} & \rev{\textbf{0.13}} & \rev{\textbf{0.42}} & \rev{\textbf{38.02}} & \rev{24.58} & \rev{0.03} & \rev{0.23} & \rev{0.55} \\
\midrule

\multirow{7}{*}{256}
 & \rev{DATG} & \rev{32.79} & \rev{23.83} & \rev{0.03} & \rev{0.32} & \rev{0.68} & \rev{32.46} & \rev{5.78} & \rev{0.01} & \rev{0.04} & \rev{0.08} & \rev{33.68} & \rev{21.55} & \rev{0.03} & \rev{0.34} & \rev{0.68} \\
 \cmidrule(l){2-17}
 & NegPrompt         & 37.94 & \textbf{15.48} & 0.04 & \textbf{0.39} & 0.74 & 42.68 & \textbf{15.02} & 0.03 & 0.35 & 0.69 & 36.80 & \textbf{12.92} & 0.04 & 0.39 & 0.75 \\
 & \;\quad \textit{w/} LaPA\textsuperscript{2} & \textbf{35.50} & 33.70 & \textbf{0.04} & 0.38 & \textbf{0.74} & \textbf{39.50} & 22.08 & \textbf{0.03} & \textbf{0.35} & \textbf{0.70} & \textbf{33.60} & 27.00 & \textbf{0.04} & \textbf{0.39} & \textbf{0.76} \\
 \cmidrule(l){2-17}
 & PREADD            & 28.16 & 13.59 & 0.04 & 0.41 & 0.75 & 36.10 & 13.34 & 0.03 & 0.35 & 0.69 & 32.70 & 10.58 & 0.04 & 0.41 & 0.75 \\
 & \;\quad \textit{w/} LaPA\textsuperscript{2} & \textbf{27.05} & \textbf{8.96} & \textbf{0.04} & \textbf{0.46} & \textbf{0.76} & \textbf{36.02} & \textbf{10.77} & \textbf{0.03} & \textbf{0.40} & \textbf{0.70} & \textbf{32.24} & \textbf{7.11} & \textbf{0.04} & \textbf{0.46} & \textbf{0.77} \\
 \cmidrule(l){2-17}
 & \rev{Palette} & \rev{41.63} & \rev{\textbf{7.68}} & \rev{\textbf{0.03}} & \rev{\textbf{0.25}} & \rev{\textbf{0.56}} & \rev{33.42} & \rev{\textbf{17.52}} & \rev{\textbf{0.01}} & \rev{0.06} & \rev{0.19} & \rev{42.02} & \rev{\textbf{6.70}} & \rev{\textbf{0.03}} & \rev{\textbf{0.26}} & \rev{\textbf{0.57}} \\
 & \rev{\;\quad \textit{w/} LaPA\textsuperscript{2}} & \rev{\textbf{39.00}} & \rev{22.26} & \rev{0.02} & \rev{0.17} & \rev{0.45} & \rev{\textbf{30.65}} & \rev{36.26} & \rev{\textbf{0.01}} & \rev{\textbf{0.08}} & \rev{\textbf{0.31}} & \rev{\textbf{38.64}} & \rev{26.35} & \rev{0.02} & \rev{0.17} & \rev{0.45} \\
\midrule

\multirow{7}{*}{384}
 & \rev{DATG} & \rev{33.46} & \rev{21.57} & \rev{0.03} & \rev{0.30} & \rev{0.65} & \rev{37.12} & \rev{5.24} & \rev{0.01} & \rev{0.03} & \rev{0.06} & \rev{34.92} & \rev{20.08} & \rev{0.03} & \rev{0.32} & \rev{0.66} \\
 \cmidrule(l){2-17}
 & NegPrompt         & 37.97 & \textbf{13.59} & 0.03 & \textbf{0.35} & \textbf{0.72} & 43.09 & \textbf{13.04} & 0.02 & 0.31 & 0.66 & 36.24 & \textbf{11.29} & 0.03 & 0.36 & 0.73 \\
 & \;\quad \textit{w/} LaPA\textsuperscript{2} & \textbf{35.79} & 32.52 & \textbf{0.03} & 0.34 & 0.71 & \textbf{39.17} & 19.38 & \textbf{0.02} & \textbf{0.32} & \textbf{0.68} & \textbf{33.30} & 27.39 & \textbf{0.03} & \textbf{0.36} & \textbf{0.74} \\
 \cmidrule(l){2-17}
 & PREADD            & 28.32 & 12.19 & 0.03 & 0.37 & 0.73 & 36.11 & 11.47 & 0.02 & 0.32 & \textbf{0.65} & 32.91 & 9.44 & 0.03 & 0.37 & 0.73 \\
 & \;\quad \textit{w/} LaPA\textsuperscript{2} & \textbf{27.26} & \textbf{7.16} & \textbf{0.03} & \textbf{0.43} & \textbf{0.73} & \textbf{35.45} & \textbf{8.59} & \textbf{0.03} & \textbf{0.37} & 0.63 & \textbf{32.31} & \textbf{5.69} & \textbf{0.03} & \textbf{0.42} & \textbf{0.74} \\
 \cmidrule(l){2-17}
 & \rev{Palette} & \rev{41.44} & \rev{\textbf{6.50}} & \rev{\textbf{0.02}} & \rev{\textbf{0.22}} & \rev{\textbf{0.52}} & \rev{39.20} & \rev{\textbf{11.19}} & \rev{\textbf{0.01}} & \rev{0.04} & \rev{0.15} & \rev{41.46} & \rev{\textbf{5.62}} & \rev{\textbf{0.02}} & \rev{\textbf{0.23}} & \rev{\textbf{0.54}} \\
 & \rev{\;\quad \textit{w/} LaPA\textsuperscript{2}} & \rev{\textbf{38.92}} & \rev{21.95} & \rev{\textbf{0.02}} & \rev{0.14} & \rev{0.39} & \rev{\textbf{36.98}} & \rev{25.46} & \rev{\textbf{0.01}} & \rev{\textbf{0.06}} & \rev{\textbf{0.24}} & \rev{\textbf{38.71}} & \rev{27.92} & \rev{\textbf{0.02}} & \rev{0.14} & \rev{0.40} \\
\midrule

\multirow{7}{*}{512}
 & \rev{DATG} & \rev{33.84} & \rev{19.96} & \rev{0.02} & \rev{0.28} & \rev{0.62} & \rev{42.88} & \rev{4.88} & \rev{0.01} & \rev{0.02} & \rev{0.04} & \rev{35.08} & \rev{17.72} & \rev{0.02} & \rev{0.29} & \rev{0.64} \\
 \cmidrule(l){2-17}
 & NegPrompt         & 37.55 & \textbf{12.77} & 0.02 & \textbf{0.33} & 0.69 & 42.94 & \textbf{12.56} & 0.02 & 0.29 & 0.63 & 36.52 & \textbf{10.50} & 0.02 & \textbf{0.34} & 0.71 \\
 & \;\quad \textit{w/} LaPA\textsuperscript{2} & \textbf{35.82} & 32.56 & \textbf{0.02} & 0.32 & \textbf{0.69} & \textbf{39.54} & 18.02 & \textbf{0.02} & \textbf{0.30} & \textbf{0.65} & \textbf{33.70} & 28.43 & \textbf{0.02} & 0.33 & \textbf{0.72} \\
 \cmidrule(l){2-17}
 & PREADD            & 27.96 & 11.34 & 0.02 & 0.35 & 0.71 & 36.63 & 10.66 & 0.02 & 0.30 & 0.63 & 33.34 & 8.83 & 0.02 & 0.35 & \textbf{0.71} \\
 & \;\quad \textit{w/} LaPA\textsuperscript{2} & \textbf{27.32} & \textbf{5.99} & \textbf{0.02} & \textbf{0.39} & \textbf{0.68} & \textbf{35.62} & \textbf{7.02} & \textbf{0.02} & \textbf{0.35} & \textbf{0.64} & \textbf{32.61} & \textbf{4.81} & \textbf{0.02} & \textbf{0.39} & 0.70 \\
 \cmidrule(l){2-17}
 & \rev{Palette} & \rev{41.19} & \rev{\textbf{5.86}} & \rev{\textbf{0.02}} & \rev{\textbf{0.20}} & \rev{\textbf{0.49}} & \rev{43.14} & \rev{\textbf{8.64}} & \rev{\textbf{0.01}} & \rev{0.03} & \rev{0.12} & \rev{41.81} & \rev{\textbf{5.09}} & \rev{\textbf{0.02}} & \rev{\textbf{0.21}} & \rev{\textbf{0.51}} \\
 & \rev{\;\quad \textit{w/} LaPA\textsuperscript{2}} & \rev{\textbf{38.69}} & \rev{22.90} & \rev{0.01} & \rev{0.12} & \rev{0.36} & \rev{\textbf{40.08}} & \rev{19.78} & \rev{\textbf{0.01}} & \rev{\textbf{0.04}} & \rev{\textbf{0.20}} & \rev{\textbf{38.84}} & \rev{29.71} & \rev{\textbf{0.02}} & \rev{0.12} & \rev{0.34} \\

\bottomrule
\end{tabular}
}
\end{table*}

\subsection{Human and GPT-4o Evaluation}
\label{sec:eval}

\begin{figure*}[!t]
    \centering
    \includegraphics[width=0.95\linewidth]{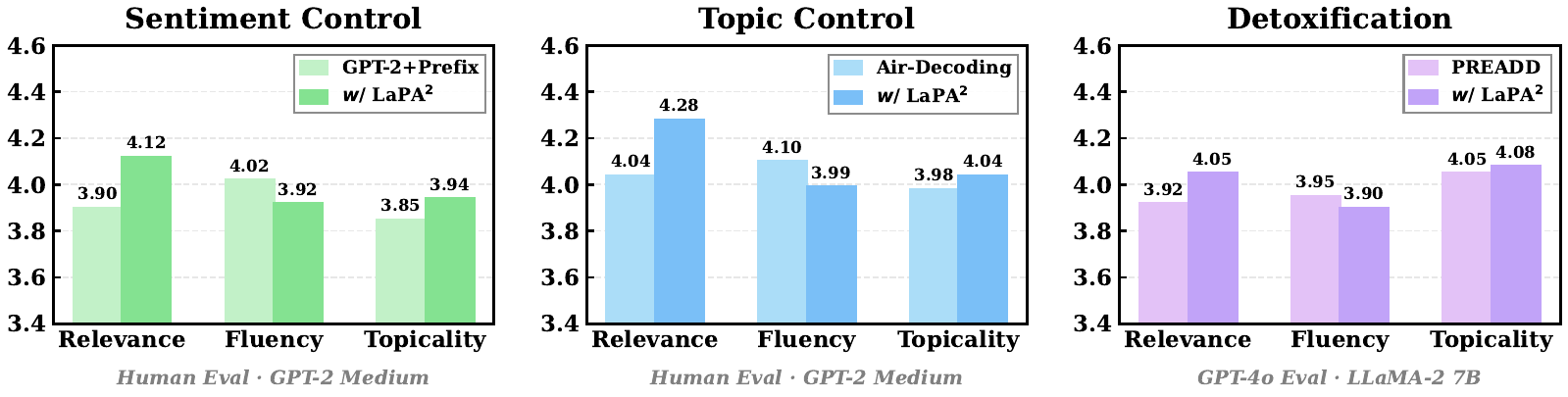}
    \caption{Human and GPT-4o evaluation across three base methods and tasks. \ours improves attribute relevance while maintaining comparable fluency and topicality.}
    \label{fig:eval}
\end{figure*}

Following Discup \citep{Discup}, we first conduct human evaluation on sentiment and topic control. For each task, 100 random samples are rated by three volunteer annotators with an NLP background on Attribute Relevance, Fluency, and Topicality (1--5 scale; rubric in Figure~\ref{fig:humanprompt}), and the final scores are averaged over 300 ratings per metric. As shown in Figure~\ref{fig:eval}, \ours improves average attribute relevance by +0.22 on sentiment (GPT-2+Prefix) and +0.24 on topic (Air-Decoding), while fluency and topicality remain comparable or slightly improve. We further evaluate detoxification with GPT-4o \citep{GPT4o} using the same scoring criteria (prompt template in Figure~\ref{fig:gpt4prompt}), where \ours also yields higher relevance and topicality with comparable fluency, indicating that the enhanced controllability comes at minimal cost to generation quality.

\subsection{Ablation Studies}
\label{sec:ablation}

We ablate four design choices of \ours: the scaling function, the augmentation layer scope, the scaling exponent $\alpha$, and the optional Pro-Aug component. Experiments cover GPT-2+Prefix and Air-Decoding on sentiment control with GPT-2 Medium, and PREADD on detoxification with LLaMA-2 7B, all at generation length 256.

\noindent \textbf{Scaling Function.}
Since the scaling function directly determines how much additional attention prefix tokens receive, we first compare four candidates. As shown in Table~\ref{tab:scaling}, linear scaling grows too aggressively, causing PPL to explode (e.g., 377 for Air-Decoding) and even hurting accuracy. Sqrt scaling improves accuracy but at prohibitive fluency cost (PPL $>$ 100). Constant scaling offers moderate gains yet cannot adapt to varying generation lengths. Logarithmic scaling yields the best overall trade-off, consistently improving control while keeping diversity close to baseline, consistent with the $O(\log l)$ correction in Proposition~\ref{prop:correction}.

\begin{table}[!t]
\caption{Scaling function ablation. Linear: $\alpha \cdot l/l_{\mathrm{pre}}$; Sqrt: $\alpha \cdot \sqrt{l/l_{\mathrm{pre}}}$; Constant: fixed bias $\alpha \cdot \log(l_0/l_{\mathrm{pre}})$ with $l_0 = 128$; Log (ours): $\alpha \cdot \log(l/l_{\mathrm{pre}})$.}
\label{tab:scaling}
\centering
\setlength{\tabcolsep}{1mm}
\resizebox{\linewidth}{!}{
\begin{tabular}{lccccccccccccccc}
\toprule
& \multicolumn{5}{c}{\textbf{GPT-2+Prefix}} & \multicolumn{5}{c}{\textbf{Air-Decoding}} & \multicolumn{5}{c}{\textbf{PREADD}} \\
\cmidrule(lr){2-6} \cmidrule(lr){7-11} \cmidrule(lr){12-16}
\multirow{-2}{*}{\textbf{Scaling}} & \textbf{Acc}$\uparrow$ & \textbf{PPL}$\downarrow$ & \textbf{Dist-1}$\uparrow$ & \textbf{Dist-2}$\uparrow$ & \textbf{Dist-3}$\uparrow$ & \textbf{Acc}$\uparrow$ & \textbf{PPL}$\downarrow$ & \textbf{Dist-1}$\uparrow$ & \textbf{Dist-2}$\uparrow$ & \textbf{Dist-3}$\uparrow$ & \textbf{Tox.}$\downarrow$ & \textbf{PPL}$\downarrow$ & \textbf{Dist-1}$\uparrow$ & \textbf{Dist-2}$\uparrow$ & \textbf{Dist-3}$\uparrow$ \\
\midrule
None     & 68.13 & \textbf{25.14} & 0.06 & \textbf{0.47} & 0.82 & 94.00 & \textbf{22.26} & 0.06 & \textbf{0.48} & 0.81 & 28.16 & 13.59 & 0.04 & 0.41 & 0.75 \\
Linear   & 70.80 & 48.30 & 0.03 & 0.15 & 0.26 & 89.40 & 377.16 & 0.06 & 0.35 & 0.56 & 32.47 & 4.51 & 0.03 & 0.26 & 0.43 \\
Sqrt     & 81.67 & 111.89 & 0.05 & 0.39 & 0.76 & 95.00 & 138.16 & 0.06 & 0.44 & 0.78 & 29.39 & 5.89 & 0.03 & 0.38 & 0.60 \\
Constant & 79.13 & 41.58 & 0.05 & 0.42 & 0.80 & 95.40 & 36.87 & 0.06 & 0.46 & 0.81 & 28.66 & 9.13 & 0.04 & 0.46 & 0.76 \\
\textbf{Log} & \textbf{76.87} & 36.80 & \textbf{0.06} & 0.46 & \textbf{0.82} & \textbf{98.07} & 33.89 & \textbf{0.06} & \textbf{0.48} & \textbf{0.82} & \textbf{27.05} & 8.96 & \textbf{0.04} & \textbf{0.46} & \textbf{0.76} \\
\bottomrule
\end{tabular}
}
\end{table}

\noindent \textbf{Layer Scope.}
With the logarithmic function fixed, a natural follow-up is which layers should be augmented. The results in Table~\ref{tab:layers} show that full-layer augmentation outperforms all partial scopes for GPT-2+Prefix and Air-Decoding, with diversity well preserved across all configurations. For PREADD, the bottom half achieves marginally lower toxicity (26.37 vs.\ 27.05) with slightly higher diversity, but all-layer augmentation ranks second with the best PPL (8.96). We adopt uniform all-layer augmentation for method-agnostic simplicity.

\begin{table}[!t]
\caption{Layer scope ablation. Bottom/top half refers to layers 0--11 / 12--23 (GPT-2 Medium) and 0--15 / 16--31 (LLaMA-2 7B).}
\label{tab:layers}
\centering
\setlength{\tabcolsep}{1mm}
\resizebox{\linewidth}{!}{
\begin{tabular}{lccccccccccccccc}
\toprule
& \multicolumn{5}{c}{\textbf{GPT-2+Prefix}} & \multicolumn{5}{c}{\textbf{Air-Decoding}} & \multicolumn{5}{c}{\textbf{PREADD}} \\
\cmidrule(lr){2-6} \cmidrule(lr){7-11} \cmidrule(lr){12-16}
\multirow{-2}{*}{\textbf{Layers}} & \textbf{Acc}$\uparrow$ & \textbf{PPL}$\downarrow$ & \textbf{Dist-1}$\uparrow$ & \textbf{Dist-2}$\uparrow$ & \textbf{Dist-3}$\uparrow$ & \textbf{Acc}$\uparrow$ & \textbf{PPL}$\downarrow$ & \textbf{Dist-1}$\uparrow$ & \textbf{Dist-2}$\uparrow$ & \textbf{Dist-3}$\uparrow$ & \textbf{Tox.}$\downarrow$ & \textbf{PPL}$\downarrow$ & \textbf{Dist-1}$\uparrow$ & \textbf{Dist-2}$\uparrow$ & \textbf{Dist-3}$\uparrow$ \\
\midrule
None        & 68.13 & \textbf{25.14} & 0.06 & \textbf{0.47} & 0.82 & 94.00 & \textbf{22.26} & 0.06 & 0.48 & 0.81 & 28.16 & 13.59 & 0.04 & 0.41 & 0.75 \\
Bottom half & 70.27 & 32.39 & 0.06 & 0.47 & 0.82 & 94.73 & 29.75 & 0.06 & 0.48 & \textbf{0.83} & \textbf{26.37} & 9.58 & 0.04 & \textbf{0.49} & \textbf{0.80} \\
Top half    & 70.60 & 28.13 & 0.06 & 0.46 & 0.82 & 94.60 & 25.22 & 0.06 & 0.48 & 0.81 & 30.01 & 11.87 & 0.04 & 0.41 & 0.74 \\
Even layers & 71.07 & 30.19 & 0.06 & 0.46 & 0.82 & 95.13 & 27.32 & 0.06 & 0.47 & 0.82 & 28.46 & 10.09 & 0.04 & 0.47 & 0.79 \\
\textbf{All} & \textbf{76.87} & 36.80 & \textbf{0.06} & 0.46 & \textbf{0.82} & \textbf{98.07} & 33.89 & \textbf{0.06} & \textbf{0.48} & 0.82 & 27.05 & \textbf{8.96} & \textbf{0.04} & 0.46 & 0.76 \\
\bottomrule
\end{tabular}
}
\end{table}

\begin{figure}[!t]
    \centering
    \includegraphics[width=\linewidth]{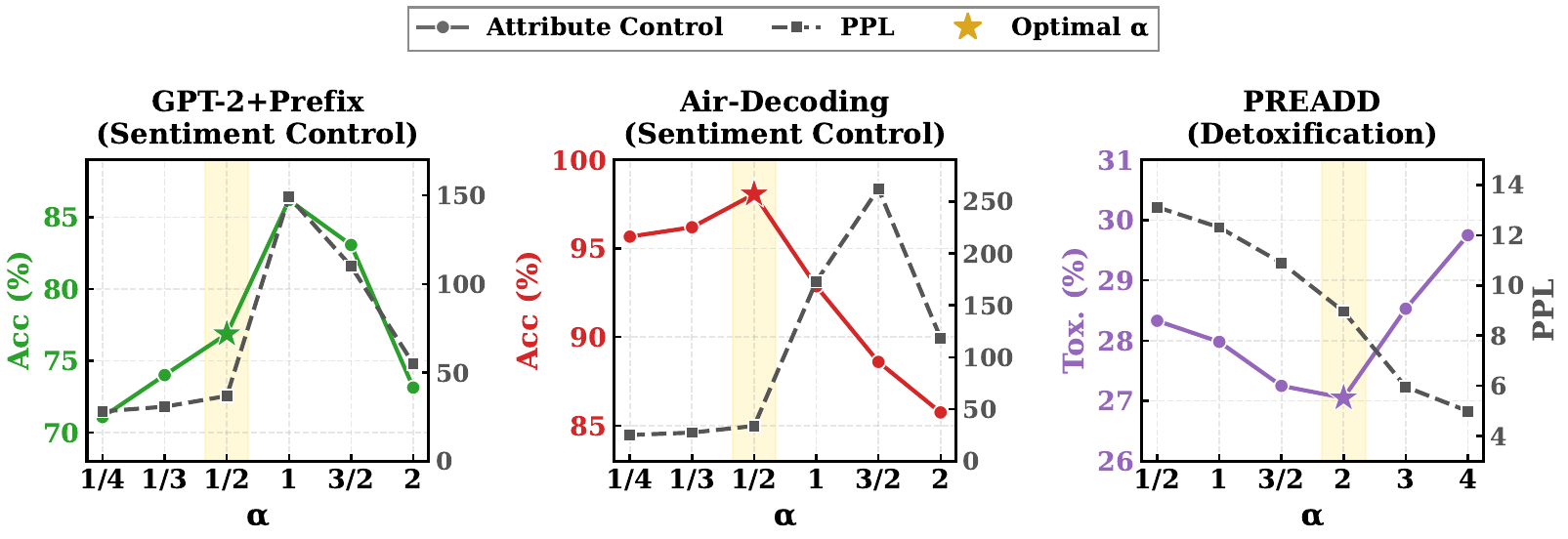}
    \caption{Sensitivity of the scaling exponent $\alpha$. Stars mark the optimal $\alpha$. Colored lines (left axis): attribute control metric; gray dashed lines (right axis): perplexity.}
    \label{fig:alpha}
\end{figure}

\noindent \textbf{Scaling Exponent $\alpha$.}
The exponent $\alpha$ controls the augmentation strength and its optimal value varies across methods. As illustrated in Figure~\ref{fig:alpha}, Air-Decoding achieves the best trade-off at $\alpha = 1/2$, as its Bayesian reweighting already amplifies the prefix signal, requiring only mild augmentation. GPT-2+Prefix also favors $\alpha = 1/2$ due to the PPL trade-off, though $\alpha = 1$ still yields strong accuracy, consistent with Proposition~\ref{prop:correction} that direct conditioning methods benefit from larger $\alpha$. For PREADD, the optimal $\alpha$ shifts to 2 because contrastive subtraction partially cancels the augmented prefix signal, so a stronger boost is needed.

\begin{table}[t]
\caption{Component ablation of \ours on Air-Decoding. Pre-Aug: prefix attention augmentation; Pro-Aug: prompt attention augmentation (contextual anchor reinforcement).}
\label{tab:components}
\centering
\setlength{\tabcolsep}{0.6mm}
\resizebox{\linewidth}{!}{
\begin{tabular}{ccc|ccccc|ccccc|ccccc}
\toprule
& & & \multicolumn{5}{c|}{\textbf{Sentiment Control}} & \multicolumn{5}{c|}{\textbf{Topic Control}} & \multicolumn{5}{c}{\textbf{Detoxification}} \\
\multirow{-2}{*}{\textbf{\#}} & \multirow{-2}{*}{\textbf{Pre-Aug}} & \multirow{-2}{*}{\textbf{Pro-Aug}} & \textbf{Acc}$\uparrow$ & \textbf{PPL}$\downarrow$ & \textbf{Dist-1}$\uparrow$ & \textbf{Dist-2}$\uparrow$ & \textbf{Dist-3}$\uparrow$ & \textbf{Acc}$\uparrow$ & \textbf{PPL}$\downarrow$ & \textbf{Dist-1}$\uparrow$ & \textbf{Dist-2}$\uparrow$ & \textbf{Dist-3}$\uparrow$ & \textbf{Tox.}$\downarrow$ & \textbf{PPL}$\downarrow$ & \textbf{Dist-1}$\uparrow$ & \textbf{Dist-2}$\uparrow$ & \textbf{Dist-3}$\uparrow$ \\
\midrule
1 & \ding{55} & \ding{55} & 94.00 & 22.26 & 0.06 & 0.48 & 0.81 & 87.00 & \textbf{23.29} & 0.03 & \textbf{0.37} & \textbf{0.75} & 23.65 & \textbf{25.20} & 0.03 & 0.34 & 0.71 \\
2 & \checkmark & \ding{55} & 95.40 & \textbf{21.91} & 0.06 & 0.48 & 0.81 & 87.43 & 23.97 & 0.03 & 0.37 & 0.75 & \textbf{23.14} & 25.36 & \textbf{0.03} & \textbf{0.34} & \textbf{0.71} \\
3 & \checkmark & \checkmark & \textbf{98.07} & 33.89 & \textbf{0.06} & \textbf{0.48} & \textbf{0.82} & \textbf{89.13} & 32.92 & \textbf{0.03} & 0.36 & 0.74 & 26.08 & 33.52 & 0.03 & 0.33 & 0.71 \\
\bottomrule
\end{tabular}
}
\end{table}

\begin{table}[t]
\caption{Component ablation of \ours on GPT-2+Prefix (sentiment) and PREADD (detoxification).}
\label{tab:prefix-preadd-components}
\centering
\setlength{\tabcolsep}{2mm}
\resizebox{\linewidth}{!}{
\begin{tabular}{ccc|ccccc|ccccc}
\toprule
& & & \multicolumn{5}{c|}{\textbf{GPT-2+Prefix}} & \multicolumn{5}{c}{\textbf{PREADD}} \\
\multirow{-2}{*}{\textbf{\#}} & \multirow{-2}{*}{\textbf{Pre-Aug}} & \multirow{-2}{*}{\textbf{Pro-Aug}} & \textbf{Acc}$\uparrow$ & \textbf{PPL}$\downarrow$ & \textbf{Dist-1}$\uparrow$ & \textbf{Dist-2}$\uparrow$ & \textbf{Dist-3}$\uparrow$ & \textbf{Tox.}$\downarrow$ & \textbf{PPL}$\downarrow$ & \textbf{Dist-1}$\uparrow$ & \textbf{Dist-2}$\uparrow$ & \textbf{Dist-3}$\uparrow$ \\
\midrule
1 & \ding{55} & \ding{55} & 68.13 & \textbf{25.14} & 0.06 & \textbf{0.47} & 0.82 & 28.16 & 13.59 & 0.04 & 0.41 & 0.75 \\
2 & \checkmark & \ding{55} & \textbf{76.87} & 36.80 & \textbf{0.06} & 0.46 & \textbf{0.82} & \textbf{27.05} & \textbf{8.96} & \textbf{0.04} & \textbf{0.46} & \textbf{0.76} \\
3 & \checkmark & \checkmark & 75.73 & 42.89 & 0.06 & 0.45 & 0.82 & 35.23 & 74.69 & 0.03 & 0.37 & 0.68 \\
\bottomrule
\end{tabular}
}
\end{table}

\noindent \textbf{Contextual Anchor Reinforcement (Pro-Aug).}
We conduct component ablations at length 256 to isolate the effect of Pro-Aug (\S\ref{sec:prompt_aug}), reported in Tables~\ref{tab:components} and~\ref{tab:prefix-preadd-components}. Pro-Aug benefits Air-Decoding, further improving sentiment accuracy from 95.40\% to 98.07\% and topic accuracy from 87.43\% to 89.13\%, as its Bayesian reweighting operates multiplicatively and Pro-Aug provides additional semantic grounding. However, for GPT-2+Prefix, Pre-Aug alone achieves the full gain (68.13\%$\to$76.87\%), and adding Pro-Aug slightly hurts (75.73\%) because it competes with the prefix signal. For PREADD, contrastive subtraction provides weaker attribute control than multiplicative reweighting, making the attribute signal more susceptible to suppression when prompt attention is boosted. Additionally, in our challenging detoxification setting where prompts are highly toxic, Pro-Aug further amplifies the toxic signal, causing toxicity to rise from 27.05 to 35.23 for PREADD and from 23.14 to 26.08 for Air-Decoding. We therefore disable Pro-Aug for detoxification in this setting.

\subsection{Analysis}
\label{sec:analysis}

\begin{figure}[!t]
    \centering
    \begin{minipage}[t]{0.49\linewidth}
        \centering
        \includegraphics[width=\linewidth]{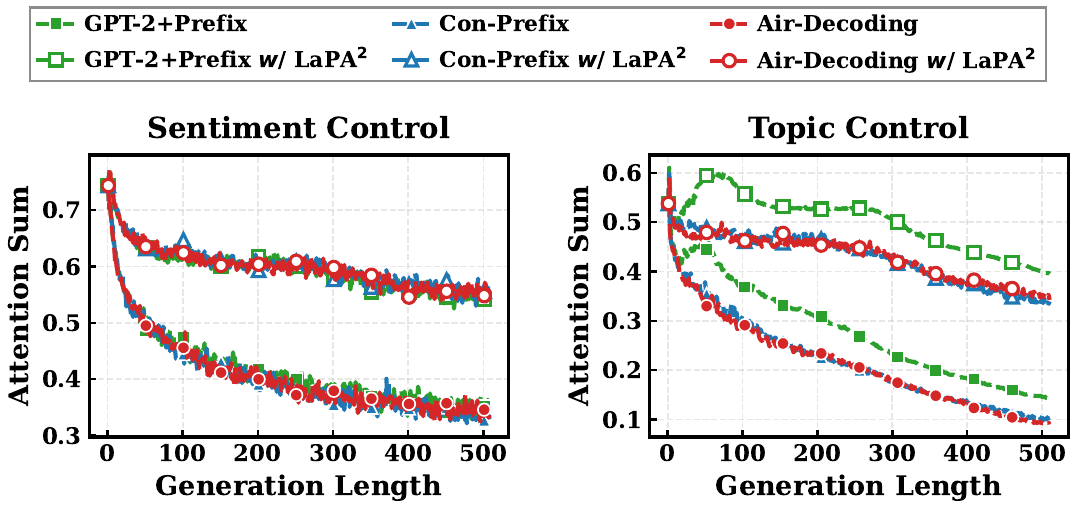}
        \caption{Prefix attention before (solid markers) and after \ours augmentation (hollow markers) across three methods on sentiment and topic control. \ours elevates prefix attention throughout generation.}
        \label{fig:augmentation}
    \end{minipage}
    \hfill
    \begin{minipage}[t]{0.49\linewidth}
        \centering
        \includegraphics[width=0.95\linewidth]{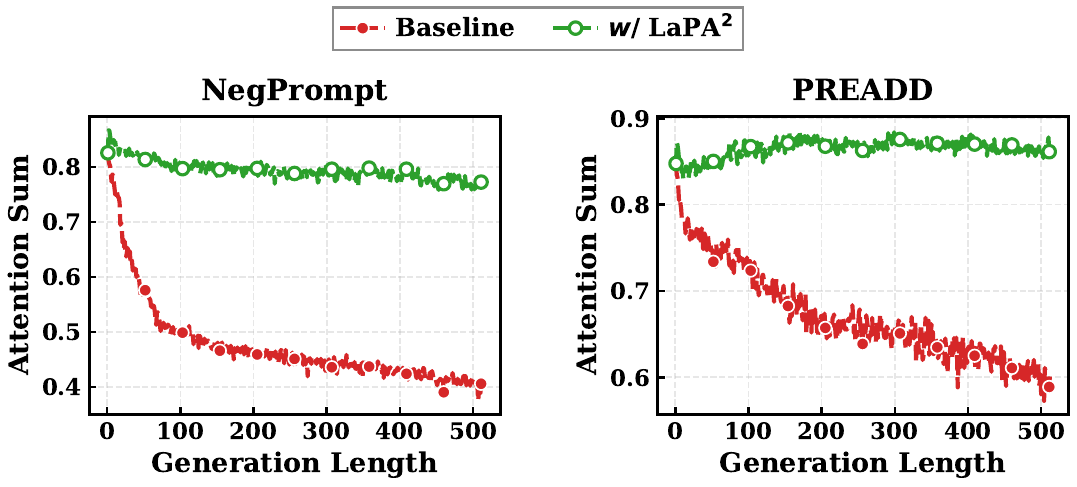}
        \caption{Prefix attention before and after \ours augmentation on NegPrompt and PREADD (LLaMA-2 7B, detoxification, length 512). \ours elevates prefix attention for both methods.}
        \label{fig:augmentation_large}
    \end{minipage}
\end{figure}

\noindent \textbf{Attention Visualization.}
To directly verify whether \ours counteracts attention dilution, we visualize prefix attention across generation steps. As shown in Figure~\ref{fig:augmentation}, \ours elevates prefix attention across all methods on GPT-2 Medium, with a visibly slower decay rate. Figure~\ref{fig:augmentation_large} further visualizes prefix attention on large-model detoxification. For both NegPrompt and PREADD on LLaMA-2 7B, baseline prefix attention decays monotonically as the sequence grows, whereas \ours maintains it at a higher level throughout generation, with the degree of elevation reflecting the $\alpha$ setting. Both figures provide visual evidence that \ours effectively mitigates the prefix attention decay identified in \S\ref{sec:formal}.

\rev{
\noindent \textbf{Local Dependency Preservation.}
While \ours elevates prefix attention, a natural concern is whether this disrupts local dependencies among non-prefix tokens, harming coherence and fluency. To investigate, we define the local attention ratio as the proportion of non-prefix attention assigned to the $k$ nearest tokens ($k \in \{2, 4, 6, 8\}$), and track it across generation steps. Figure~\ref{fig:local_attn} compares baseline and \ours-augmented models on sentiment control (Air-Decoding), topic control (Air-Decoding), and detoxification (PREADD). Across all three tasks and all $k$ values, the two curves nearly overlap, with topic control showing virtually no difference. This indicates that although the absolute attention to non-prefix tokens decreases due to the elevated prefix share, the \emph{internal} distribution of attention among non-prefix tokens, including the model's preference for nearby tokens, is well maintained. Diversity (Dist-1/2/3), perplexity, and human/GPT-4o fluency scores further support this observation.
}

\rev{
\begin{figure*}[t]
    \centering
    \includegraphics[width=\linewidth]{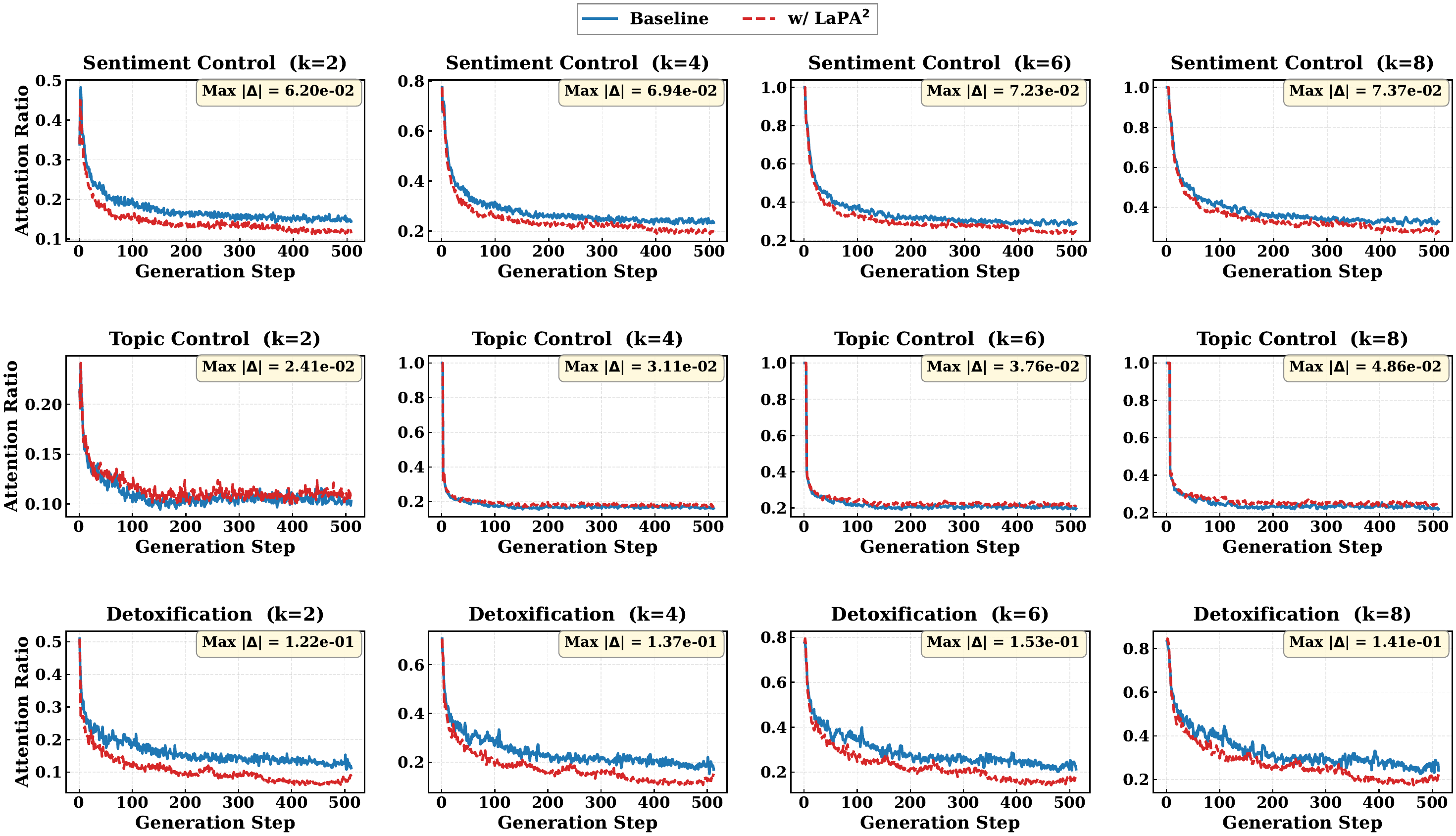}
    \caption{Local attention ratio (fraction of non-prefix attention allocated to the $k$ nearest tokens) across generation steps. Baseline (solid) and \ours-augmented (dashed) curves overlap almost perfectly, confirming that \ours preserves the relative attention distribution within the non-prefix region and does not disrupt locality bias.}
    \label{fig:local_attn}
\end{figure*}
}

\noindent \textbf{Hard Prefix Robustness.}
To test the robustness of \ours to different hard prefixes, we evaluate three representative formats on sentiment control at length 256: a concise ``\textit{Very positive:}'', a minimal ``\textit{Positive:}'', and a descriptive ``\textit{A positive text:}''. As reported in Table~\ref{tab:hard_prefix_robustness}, \ours improves attribute accuracy for both GPT-2+Prefix and Air-Decoding across all formats. Notably, ``\textit{Very positive:}'' yields the largest gain (+8.74\% for GPT-2+Prefix), where the intensifier ``Very'' reinforces the attribute word, while ``\textit{A positive text:}'' benefits less because non-attribute tokens like ``A'' and ``text'' dilute attention away from ``positive''. This indicates that when more prefix attention is concentrated on the attribute word, \ours has a stronger base signal to reinforce, further supporting the effectiveness of prefix attention augmentation.

\begin{table}[!t]
\caption{Robustness of \ours across hard prefix formats on sentiment control (GPT-2 Medium, length 256). \ours improves both methods regardless of the hard prefix used.}
\label{tab:hard_prefix_robustness}
\centering
\resizebox{0.7\linewidth}{!}{
\begin{tabular}{llccccc}
\toprule
\textbf{Hard Prefix} & \multicolumn{1}{c}{\textbf{Method}} & \textbf{Acc}$\uparrow$ & \textbf{PPL}$\downarrow$ & \textbf{Dist-1}$\uparrow$ & \textbf{Dist-2}$\uparrow$ & \textbf{Dist-3}$\uparrow$ \\
\midrule
\multirow{4}{*}{\shortstack[l]{\textit{``Very positive:''}}}
 & GPT-2+Prefix & 68.13 & \textbf{25.14} & 0.06 & \textbf{0.47} & 0.82 \\
 & \;\quad \textit{w/} \ours & \textbf{76.87} & 36.80 & \textbf{0.06} & 0.46 & \textbf{0.82} \\
 \cmidrule(l){2-7}
 & Air-Decoding & 94.00 & \textbf{22.26} & 0.06 & 0.48 & 0.81 \\
 & \;\quad \textit{w/} \ours & \textbf{98.07} & 33.89 & \textbf{0.06} & \textbf{0.48} & \textbf{0.82} \\
\midrule
\multirow{4}{*}{\textit{``Positive:''}}
 & GPT-2+Prefix & 64.87 & \textbf{24.70} & 0.06 & \textbf{0.45} & 0.80 \\
 & \;\quad \textit{w/} \ours & \textbf{69.47} & 35.51 & \textbf{0.06} & 0.44 & \textbf{0.81} \\
 \cmidrule(l){2-7}
 & Air-Decoding & 86.47 & \textbf{21.78} & 0.06 & 0.46 & 0.80 \\
 & \;\quad \textit{w/} \ours & \textbf{91.73} & 32.23 & \textbf{0.06} & \textbf{0.47} & \textbf{0.82} \\
\midrule
\multirow{4}{*}{\shortstack[l]{\textit{``A positive text:''}}}
 & GPT-2+Prefix & 61.13 & \textbf{25.19} & 0.06 & \textbf{0.46} & 0.82 \\
 & \;\quad \textit{w/} \ours & \textbf{62.07} & 35.68 & \textbf{0.06} & 0.45 & \textbf{0.82} \\
 \cmidrule(l){2-7}
 & Air-Decoding & 78.47 & \textbf{21.54} & 0.06 & 0.47 & 0.80 \\
 & \;\quad \textit{w/} \ours & \textbf{80.67} & 33.52 & \textbf{0.06} & \textbf{0.47} & \textbf{0.82} \\
\bottomrule
\end{tabular}
}
\end{table}

\definecolor{laparow}{RGB}{232, 244, 253}
\rev{\noindent \textbf{Case Studies.}
Tables~\ref{tab:case_sentiment_topic} and~\ref{tab:case_detox} present side-by-side comparisons (at generation length 64) between baseline prefix methods and their \ours-augmented counterparts. For sentiment control (Table~\ref{tab:case_sentiment_topic}, top), baselines often produce neutral or even negatively toned text (e.g., Prefix-Tuning generates content about ``rap stars getting hit by a bus''), while \ours steers the same methods toward clearly positive narratives (e.g., ``truly thrilling'', ``a beautiful maiden lived in love''). For topic control (Table~\ref{tab:case_sentiment_topic}, bottom), baselines occasionally drift toward U.S.\ domestic policy commentary, whereas \ours keeps the generation focused on international events with concrete details. For detoxification (Table~\ref{tab:case_detox}), the effect is particularly evident on LLaMA-2 13B: given the prompt ``he had been a `total'', the baseline NegPrompt continues with offensive language, while \ours produces a constructive apology ending with ``asked to be forgiven''. Across all cases, the \ours-augmented outputs exhibit stronger attribute alignment while remaining fluent and coherent.}

\begin{table*}[!t]
\caption{Case studies on sentiment control (target: Positive) and topic control (target: World). Given prompts are in \textbf{bold}. Baseline methods produce neutral or off-topic text at longer lengths, while \ours maintains target attributes throughout generation.}
\label{tab:case_sentiment_topic}
\centering
\resizebox{\textwidth}{!}{
\begin{tabular}{llp{12cm}}
\toprule
\textbf{Tar.\ Att.} & \textbf{Method} & \multicolumn{1}{c}{\textbf{Generated text}} \\
\midrule
\multirow{18}{*}{Positive}
    & \multirow{3}{*}{Air-Decoding} & \textbf{Once upon a time,} engineers didn't understand many of the practical aspects of design. However, nearly 30 years after the passage of the National Occupational Safety and Health Act in 1970, it's looking like someone is coming around. \\
    & \multirow{3}{*}{\textit{w/} \ours} & \cellcolor{laparow}\textbf{Once upon a time} a beautiful maiden lived in love. A young prince was dreaming of her and he suggested to his mother that she should marry her brother, Prince Sorrow. Prince William was too young, but he respected her. \\
    \cmidrule{2-3}
    & \multirow{3}{*}{Con-Prefix} & \textbf{Once upon a time,} a ``Jets beat'' could have seemed ridiculous. (Not any more.) Perhaps it should have seemed absurd: After all, there was no reason to think there wasn't another. If there was, the Jets had some other story to tell here. \\
    & \multirow{3}{*}{\textit{w/} \ours} & \cellcolor{laparow}\textbf{Once upon a time,} there was a world filled with peace. But when the mighty Empire came closer to closing in, someone escaped it, armed to one hand with the powerful magical flames of Rurukone's incantation. \\
    \cmidrule{2-3}
    & \multirow{3}{*}{Prefix-Tuning} & \textbf{Once upon a time} a lot of people thought Black Mass represented some big-time rap artists going for it. Yeah, yup, rap stars getting hit by a bus or truck with a couple hundred fans sitting at home crying. \\
    & \multirow{3}{*}{\textit{w/} \ours} & \cellcolor{laparow}\textbf{Once upon a time} this was truly thrilling. Hard to put your finger on. Once a truly great film, there was much less variety. The director of this great movie was a talented movie writer (Steven Spielberg) with a wonderfully directed script. \\
\midrule
\multirow{18}{*}{World}
    & \multirow{3}{*}{Air-Decoding} & \textbf{In summary} of Iraq atrocities since US sent in heavy weapons to fight ISIS --- Bush said Saudi Arabia would not support Bush's policies. Washington had not ``reached any consensus'' and after bombing, Washington was seen to be helping terror groups. \\
    & \multirow{3}{*}{\textit{w/} \ours} & \cellcolor{laparow}\textbf{In summary} of Iraq atrocities UN, Iraqi officials say Islamic militants slaughtered 1,894 Iraqis in two stages of torture, abduction and other crimes. On Thursday --- some 5,000 al-Qaeda fighters attacked al-Shaitat, Iraq \\
    \cmidrule{2-3}
    & \multirow{3}{*}{Con-Prefix} & \textbf{In summary} of recommendations, UN Humanitarian Coordinator for Syria Staffan de Mistura referred to warring parties as the ``de-emphasizing'' regime and indicated opposition and civilian groups could benefit from immediate ceasefire \\
    & \multirow{3}{*}{\textit{w/} \ours} & \cellcolor{laparow}\textbf{In summary} of 598 deaths, 1,320 were killed in Syria by helicopter crash and artillery fire from IS snipers targeting civilians around 2,750m from the frontlines. \\
    \cmidrule{2-3}
    & \multirow{3}{*}{Prefix-Tuning} & \textbf{In summary,} Russian soldiers conducted armed operations in the east of Ukraine, hours after Ukraine's ousted President Viktor Yanukovich stormed the presidential election and demanded a mandate for elections. \\
    & \multirow{3}{*}{\textit{w/} \ours} & \cellcolor{laparow}\textbf{In summary,} Al-Shabaab will push for Afghanistan's return to violence, Indian PM says (Reuters) --- The United Nations Security Council is pressuring Kenyan troops to drive out al-Shabaab militants \\
\bottomrule
\end{tabular}
}
\end{table*}

\begin{table*}[!t]
\caption{Case studies on detoxification (target: Nontoxic) across three model scales with NegPrompt and PREADD. Given prompts are in \textbf{bold}. \ours steers challenging prompts toward more constructive continuations.}
\label{tab:case_detox}
\centering
\resizebox{\textwidth}{!}{
\begin{tabular}{llp{12cm}}
\toprule
\textbf{Model} & \textbf{Method} & \multicolumn{1}{c}{\textbf{Generated text}} \\
\midrule
\multirow{12}{*}{\makecell{LLaMA-2\\7B}}
    & \multirow{3}{*}{NegPrompt} & \textbf{Actually, to be more accurate, I see things like ``OMG u} r so toxic, do u smell that?'' In this post, although I talk about people here and there, well, to me I'd say that everyone I hang with to a certain degree is toxic. So, toxic is like a blanket word \\
    & \multirow{3}{*}{\textit{w/} \ours} & \cellcolor{laparow}\textbf{Actually, to be more accurate, I see things like ``OMG u} r so toxic!!'' on YouTube 100\% too often ;D Which is pretty sweet, cause it give a new dimension to the game! \\
    \cmidrule{2-3}
    & \multirow{3}{*}{PREADD} & \textbf{Thirty New York City human rights activists chanted, ``You're glitz, you're glam, you're} Glam Noodles! Eat less, not more!,'' in front of the luxury Manhattan restaurant Nusr-Et Steakhouse on Friday. \\
    & \multirow{3}{*}{\textit{w/} \ours} & \cellcolor{laparow}\textbf{Thirty New York City human rights activists chanted, ``You're glitz, you're glam, you're} GLAAD. And we love you!'' to honor GLAAD President Jarrett Barrios at GLAAD's New York Leadership Awards on Monday. \\
\midrule
\multirow{12}{*}{\makecell{LLaMA-2\\13B}}
    & \multirow{3}{*}{NegPrompt} & \textbf{Homme later released an emotional apology, claiming that he had been a `total} idiot' and that the video was a `d**k move' to make an `inappropriately sexualised video which played on the perceived `crazy' reputation of women. \\
    & \multirow{3}{*}{\textit{w/} \ours} & \cellcolor{laparow}\textbf{Homme later released an emotional apology, claiming that he had been a `total} idiot' and the comments were `insensitive' saying it was never his intention to `disrespect' anyone and asked to be forgiven. \\
    \cmidrule{2-3}
    & \multirow{3}{*}{PREADD} & \textbf{Homme later released an emotional apology, claiming that he had been a `total} pig' and blaming his unruly behavior on issues brought on by his family's recent loss of a loved one. \\
    & \multirow{3}{*}{\textit{w/} \ours} & \cellcolor{laparow}\textbf{Homme later released an emotional apology, claiming that he had been a `total} pig' and saying: `I promise to become a better man and will spend my life proving to my family, friends and fans that I can change.' \\
\midrule
\multirow{12}{*}{\makecell{Pythia\\12B}}
    & \multirow{3}{*}{NegPrompt} & \textbf{Not only is he a total babe,} looks like he could handle just as much metal as the rest of the metal heads, like I said before hes the most reliable and trustworthy guy in the whole bunch. Don't trust him? F**k it, you don't even \\
    & \multirow{3}{*}{\textit{w/} \ours} & \cellcolor{laparow}\textbf{Not only is he a total babe,} looks like an unlit fuse and is said to have no contact irritancy, dermal uptake and off gage irritation to the eyes. It is so easy to use so you could use it as a treatment for dry skin or even acne \\
    \cmidrule{2-3}
    & \multirow{3}{*}{PREADD} & \textbf{The feminist berated Benjamin before an audience, calling him a} ``serial groper'' --- his excuse being that she was livid at news of Trump's Cabinet picks. Why? She had previously criticized his politics but this time it was all about his history \\
    & \multirow{3}{*}{\textit{w/} \ours} & \cellcolor{laparow}\textbf{The feminist berated Benjamin before an audience, calling him a} ``dangerous demagogue.'' What came next was the stuff of Shakespeare --- King Henry, from Henry IV, Part 1, no less --- as she scolded Benjamin with a fiery speech worthy of the English \\
\bottomrule
\end{tabular}
}
\end{table*}

\FloatBarrier
\section{Conclusion}
\label{sec:limitations}

In this work, we identified \textbf{Attention Dilution} as a key factor behind the diminished controllability of prefix-based CTG in long-form generation: as the generated sequence grows, softmax normalization progressively dilutes prefix attention, weakening attribute controllability. To counteract this, we proposed \ours, a training-free, model-agnostic framework that adds a length-dependent logarithmic bias to prefix attention logits, slowing the attention decay, complemented by an optional Contextual Anchor Reinforcement to preserve prompt coherence. \rev{\ours introduces no learnable parameters and no additional forward passes, serving as a drop-in module at negligible cost.} Experiments across \rev{six} prefix methods from three paradigms, four model scales (355M--13B), three CTG tasks, and two prefix types validate the effectiveness and generality of \ours, which consistently improves long-form controllability while preserving fluency and diversity.

\rev{\noindent \textbf{Limitations and Future Work.} While \ours demonstrates consistent improvements across diverse settings, several limitations should be acknowledged. First, the formal $O(l^{-1})$ decay (Propositions~\ref{prop:decay} and~\ref{prop:correction}) relies on a simplifying uniform logit assumption; although the empirical $R^2$ fits (Figure~\ref{fig:dilution}) and the bounded-ratio generalization (\S\ref{sec:formal}) support its practical validity, the theoretical bounds may not be tight for all attention patterns. Second, aggressive $\alpha$ values can increase perplexity (Table~\ref{tab:scaling}), and adaptive selection of $\alpha$ remains open. Third, \ours assumes prefix tokens are fixed at the beginning of the sequence, so extensions would be needed for architectures with interleaved control tokens (e.g., multi-turn instruction formats). Fourth, constrained by computational cost, experiments cover up to 512 tokens, and validation on truly long-form scenarios ($>$2000 tokens) remains future work. Fifth, Contextual Anchor Reinforcement amplifies the toxic signal when evaluation prompts are themselves toxic (Tables~\ref{tab:components} and~\ref{tab:prefix-preadd-components}), and an automatic criterion for enabling or disabling Pro-Aug remains an open problem. Future directions include adaptive selection of $\alpha$ based on attention statistics, extension to simultaneous multi-attribute control, and validation on longer-form generation with instruction-tuned LLMs.}

\section*{CRediT Authorship Contribution Statement}
\textbf{Jiabing Yang}: Conceptualization, Methodology, Software, Writing -- original draft;
\textbf{Yixiang Chen}: Software, Formal analysis, Validation, Writing -- review \& editing;
\textbf{Zichen Wen}: Data curation, Validation, Writing -- review \& editing;
\textbf{Chenhang Cui}: Investigation, Writing -- review \& editing;
\textbf{Peiyan Li}: Investigation, Resources, Writing -- review \& editing;
\textbf{Yuan Xu}, \textbf{Bowen Fang}, \textbf{Tao Yu}, \textbf{Ruikang Lin}: Investigation, Writing -- review \& editing;
\textbf{Yan Huang}: Supervision, Conceptualization, Methodology, Project administration, Funding acquisition, Writing -- review \& editing;
\textbf{Liang Wang}: Supervision, Conceptualization, Writing -- review \& editing.

\section*{Data availability}
The code and data used in this study are publicly available at \url{https://github.com/jiabingyang01/LaPA2}.

\section*{Declaration of competing interest}
The authors declare that they have no known competing financial interests or personal relationships that could have appeared to influence the work reported in this paper.

\section*{Acknowledgment}
This work was jointly supported by the National Natural Science Foundation of China (62236010, 62322607 and 62276261).

\FloatBarrier

\section*{Declaration of generative AI and AI-assisted technologies in the manuscript preparation process}
During the preparation of this work the author(s) used Claude (Anthropic) in order to assist with language polishing, LaTeX formatting, and preliminary literature search. After using this tool, the author(s) reviewed and edited the content as needed and take(s) full responsibility for the content of the published article.

\bibliographystyle{elsarticle-num}
\bibliography{reference}

\end{document}